\documentclass[pdflatex,sn-mathphys]{sn-jnl}

\usepackage{mathtools}
\usepackage{amsmath}
\usepackage{amsfonts}
\usepackage{hyperref}
\usepackage{amssymb}
\usepackage[numbers,sort&compress]{natbib}
\usepackage[mathscr]{euscript}

\theoremstyle{thmstyleone}%
\theoremstyle{thmstyletwo}%

\theoremstyle{thmstylethree}%

\begin{document}

\title[StressVal]{Towards certification: A complete statistical validation pipeline for supervised learning in industry}

\author[1]{Lucas Lacasa}
\author[2]{Abel Pardo}
\author[3,4]{Pablo Arbelo}
\author[3]{Miguel Sánchez} 
\author[3]{Pablo Yeste}
\author[3]{Noelia  Bascones}
\author[3]{Alejandro Martínez-Cava}
\author[3]{Gonzalo Rubio}
\author[3]{Ignacio Gómez}
\author[3,5]{Eusebio Valero}
\author[3,5]{Javier de Vicente}

\affil[1]{\orgdiv{Institute for Cross-Disciplinary Physics and Complex Systems (IFISC)}, \orgname{CSIC-UIB}, \orgaddress{\city{Palma de Mallorca}, \country{Spain}}}
\affil[2]{\orgdiv{Department of Aerostructures, Airbus},  \orgaddress{Paseo John Lennon 28906 Getafe \country{Spain}}}
\affil[3]{\orgdiv{Department of Applied Mathematics and NUMATH group}, \orgname{ETSI Aeronáutica y del Espacio, Universidad Politécnica de Madrid}, \orgaddress{\city{Madrid}, \country{Spain}}}
\affil[4]{\orgdiv{European Space Agency (ESA)},  \orgaddress{\city{Cologne}, \country{Germany}}}
\affil[5]{\orgdiv{Center for Computational Simulation}, Universidad Politécnica de Madrid, Campus de Montegancedo,
Boadilla del Monte,Madrid, Spain}


\abstract{Methods of Machine and Deep Learning are gradually being integrated into industrial operations, albeit at different speeds for different types of industries. The aerospace and aeronautical industries have recently developed a roadmap for concepts of design assurance and integration of neural network-related technologies in the aeronautical sector. 
This paper aims to contribute to this paradigm of AI-based certification in the context of supervised learning, by outlining a complete validation pipeline that integrates deep learning, optimization and statistical methods. This pipeline is composed by a directed graphical model of ten steps. Each of these steps is addressed by a merging key concepts from different contributing disciplines (from machine learning or optimization to statistics) and adapting them to an industrial scenario, as well as by developing computationally efficient algorithmic solutions. 
We illustrate the application of this pipeline in a realistic supervised problem arising in aerostructural design: predicting the likelikood of different stress-related failure modes during different airflight maneuvers based on a (large) set of features characterising the aircraft internal loads and geometric parameters. 
}




\maketitle
\section{Introduction}
\label{sec:intro}
The field of Machine Learning (ML) \cite{DL,ML} and its broad spectrum of applications has revolutionized a plethora of technological industries in recent years ranging from the energy sector or material sciences to telecommunications, finance or consumer goods, to cite some \cite{ML_app}. In the context of aeronautical engineering and aerospace technologies, the field has embraced ML tools only in recent years, and impact is growing at a rapid pace, ranging from general-purpose ML-based fluid mechanics \cite{ML_industries_0, ML_industries_1,ML_industries_2}, aeroacoustics \cite{ML_aerospace3}, wind turbines \cite{sole} or aerostructures \cite{ML_aerospace4} (including prediction of landing gear loads \cite{javi2}) to flight trajectories optimization \cite{pablo6} or enhancing predictive maintenance \cite{pablo7,pablo8}: see  the recent and illuminating reviews \cite{ML_aerospace1, ML_aerospace2} and references therein.\\ 
Interestingly, the integration of ML-related tools and ideas in the aeronautical and aerospace industries is still in its infancy. 
Part of the reason is that any new technology has a necessary adoption curve \cite{adoption, adoption2}, and the fact that ML-solutions require expert knowledge at the crossroads of computer science and statistics --and a sophisticated operationalization infrastructure (MLOps) \cite{mlops} -- does not facilitate this adoption. However, a deeper reason is probably impeding faster adoption: while ML-technologies promise high performance and reduction in development and operating costs \cite{costs} (e.g. by reducing costs related to expensive and lengthy wind tunnel experiments and numerical simulations), ensuring adequate safety remains paramount in aeronautical industries, and ML-based tools are often seen as sophisticated black-boxes that suffer from low degree of trustability, and thus difficult to validate their safety. Therefore, air safety authorities demand rigorous validation and verification processes for these models, and industry leaders have started to propose guidelines and a roadmap on concepts of design assurance for neural network-related technologies \cite{codann1, codann2, codann3}. 
However, only very recently industry has started to embrace the complexities of certifying ML models \cite{pablo1,pablo2,pablo3,pablo4,pablo5}, prompting the initiation of
discussions around the development of guidelines and a roadmap for design assurance, especially
concerning network-related technologies. This pressing need underscores the imperative
for collaborative efforts within the industry to establish robust validation frameworks that not
only meet regulatory standards but also address the evolving challenges posed by ML integration. This has indeed been well understood and undertaken by Airbus who has established an internal working group on verification and validation of surrogate models in the frame of loads and stress domains. The collaboration between Airbus and UPM highlighted in this paper happened in the context of this working group. 

\medskip
\noindent {This paper presents a comprehensive validation pipeline with the rigor required for contributing to industrial certification of ML-based solutions in the aeronautical industry. Our framework leverages established concepts from machine and deep learning, integrates sophisticated statistical methods, and introduces novel algorithms to enable a thorough --yet computationally efficient-- validation of both performance and safety. Each step of the framework is demonstrated using realistic data. The versatility of our framework has been probed by showcasing its application in diverse settings, ranging from predicting internal loads obtained from FEM simulations using a neural network trained on external loads \cite{stress3}, to verifying the decisions of a model assessing the need for aircraft inspections after hard landings. As a case study and in order to illustrate each part of the framework, in this paper we mainly consider a paradigmatic case of industrial importance in aeronautical stress engineering for which a machine learning certification is needed: the prediction of reserve factors determining the likelihood of different stress-related failure modes based on the aerodynamic loads induced by various flight maneuvers.\\
This work aims to contribute to the broader effort within the aeronautical industry to replace traditional tools like wind tunnel experiments and numerical simulations with faster and more versatile surrogate models. 
}

\medskip
\noindent The rest of the paper goes as follows. Section \ref{sec:notation} depicts some mathematical notation and general guidelines on the training algorithms. In section \ref{sec:industrial_problem} we provide some additional details of the industrial problem we will use as an illustration. The bulk of results is presented in section \ref{sec:pipeline}, where we present the pipeline: a directed graph of ten conceptual steps which we thoroughly describe --both conceptually, mathematically and algorithmically-- and illustrate. Finally, in section \ref{sec:discussion} we conclude.

\section{Mathematical notation and some general reminder on the training algorithms}
\label{sec:notation}
Many of the ML optimization problems that emerge in aerospace industry are in essence regression problems \cite{ML_aerospace1, ML_aerospace2}. Different machine learning models are available to attack supervised learning problems, ranging from classical methods such as linear regression, k-nearest neighbors or random forests to the family of
artificial neural networks (ANN) \cite{DL,ML}. Formally, despite the specific approach the task is always to find the explicit shape of a nonlinear function $\mathcal F()$ that takes a vector of features ${\bf X} \in S_X$ and outputs a vector of predictions ${\bf \tilde Y} \in S_{\tilde{Y}}$. $S_X$ and $S_{\tilde{Y}}$ are the domain and co-domain sets of the function. They are traditionally called the feature set and the prediction set, and usually we have $S_X \subset \mathbb{R}^n, \ S_{\tilde{Y}} \subset \mathbb{R}^m$ (for regression), usually (but not always) with $m\ll n$. $n$ is called {\it the number of features} or the {\it dimensionality of the ambient space}, whereas for regression $m$ is called the number of (dependent) predictions or the number of output variables.\\
In the rest of this work we will focus on solving regression problems via ANNs, although most of the discussion and validation methods remain valid for other optimization frameworks. For feed-forward (deep) networks \cite{DL}, $\mathcal F()$ consists of successive compositions of (i) affine combinations of the elements of $\bf X$, followed by (ii) a nonlinear transformation (usually a rectified linear unit or ReLu) (in the $j$-th composition, the affine combination is not of the elements of $\bf X$, but of the elements of the output of the $(j-1)$-th composition). For other architectures, the formal shape of $\mathcal F()$ is different. In any case, once the ANN architecture is fixed, essentially we have a fully parametrized function $\mathcal F()$, and thus finding the optimal shape of $\mathcal F()$ is equivalent to finding the optimal assignment to the set of parameters $\bf W$. Formally, we thus have that the function $\mathcal F$ is such that 
$$\mathcal{F}:S_X\to S_{\tilde{Y}}, {\bf {\tilde{Y}}}={\mathcal F}(\bf X;\bf W).$$
In order to find the optimal assignment of $\bf W$, as in any regression problem, we initially need to have access to a large set of duples $(\bf X, \bf Y)$. Observe here the lack of tilde in $\bf Y$, as these are not predictions of the nonlinear function, but the so-called {\it ground true} values associated to $\bf X$. The full dataset of such points is called $$\text{Full}=\{({\bf X}(i), {\bf Y}(i))\}_{i=1}^N$$
In the most basic setting, one usually proceeds to partition Full into a training set and a test set `Test', such that
\begin{eqnarray}
    &&\text{Train} \subset \text{Full}, \ \text{Test} \subset \text{Full},\ \text{Full} = \text{Train} \cup \text{Test}, \\ \nonumber  &&\text{Train} \cap \text{Test} =\emptyset, \ \vert\text{Train}\vert> \vert\text{Test}\vert    
\end{eqnarray}
The specific way in which this partition is performed is summarized into what is called the training-test (TT) split (see section \ref{sec:split} for details).\\ 
Then, the loss function is formally defined as 
\begin{equation}
  \mathcal{L}(\bf W)=\text{E}(\bf W) + \lambda \text{R}(\bf W), 
  \label{eq:loss}
\end{equation}
where 
\begin{itemize}
    \item $\text{E}()$ is an error function which accumulates the mismatch between the function predictions $\bf \tilde{Y}$ and the ground true values $\bf Y$ over all samples. Depending on the way the mismatch is explicitly computed, we have different error functions, e.g. for the Mean Square Error
    \begin{equation}
    \label{eq:Error_MSE}
        \text{E(\bf W)} = \frac{1}{N_{\text{Train}}}\sum_{i=1}^{N_{\text{Train}}} ({\bf Y}(i) - {\mathcal F}[{\bf X}(i);\bf W])^2
    \end{equation}
    whereas for the Mean Absolute Error
        \begin{equation}
    \label{eq:Error_MAE}
        \text{E(\bf W)} = \frac{1}{N_{\text{Train}}}\sum_{i=1}^{N_{\text{Train}}} \vert {\bf Y}(i) - {\mathcal F}[{\bf X}(i);\bf W]\vert
    \end{equation}
    \item $\text{R}()$ is a regularization function. There are many different possibilities, depending on the role one wants the regularizer to play, a standard choice to prevent overfitting is using $L_2$ (Tikhonov) regularization 
    $$ \text{R}(\bf W) = \vert \vert \bf W \vert \vert_2, $$
    whereas if one wants to condition the search on sparse networks, then $L_1$ regularization is usually preferred.
\end{itemize}
The training step \cite{DL} aims to find the values of $\bf W$ that minimize the loss function. Along training, the network performs a trajectory in graph space \cite{Lucas1, Lucas2, Lucas3}. The trained network will have parameters $\bf W^*$, where
$${\bf W}^*= \text{argmin}\{\mathcal{L}(\bf W)\}$$
A classical algorithm to find ${\bf W}^*$ is to employ gradient descent, that from an initial condition, iteratively searches in the direction of largest gradient:
\begin{eqnarray}
    {\bf W}(k+1) ={\bf W}(k) - \eta \nabla_{\bf W} \mathcal{L}(\bf W),
    \label{eq:GD}
\end{eqnarray}

where $\eta$ is the learning rate and differentiation is implemented via backpropagation. In practice, one standard choice for training is to use stochastic or mini-batch gradient descent (where instead of the full training set, each iteration of Eq.\ref{eq:GD} does not make use of the full error term, but only batches randomly sampled from the training set (i.e. in Eq.\ref{eq:Error_MSE} the sum is over a handful of samples, called a batch)). It is also customary to use varying learning rates (e.g. schedules or adaptive methods), such as in the standard Adam optimizer \cite{DL}.

\section{The industrial problem}
\label{sec:industrial_problem}
In order to illustrate each of the steps of the methodological framework we will introduce in the next section, we shall consider a realistic case of industrial interest in the aeronautic field: that of leveraging supervised learning methods in aircraft stress engineering.\\
 A typical semi-monoque fuselage structure consists on longitudinal stringers and skin panels, withstanding axial, hoop and shear loads as well as pressure and temperature. To size those stiffened panels is required to study diverse failure modes related to strengths and buckling \cite{stress, stress2,stress3} . For instance, forced crippling occurs when the shear buckles in the panel skin force the attached stiffener flanges to deform out-of-plane. In the task under analysis, we identify and label a total of $m=6$ failure modes:
Forced Crippling, Column Buckling, In Plane, Net Tension, Pure Compression, and Shear Panel Failure.
The task here is to train a deep ANN to accurately
predict the {\it reserve factors} (RF) that quantify the failure likelihood of these failure modes: ${\bf Y}=(y_1,y_2,\dots,y_6)$ where $y_j$ is the reserve factor of the $j$-th failure mode, and $y_j \in [0,5]$, where $0$ means extreme risk of failure and $5$ means extremely low risk (in logarithmic scale).\\
We want the ANN to be able to accurately predict ${\bf Y}$ for different regions of the aircraft (at specific combinations of frames and stringers), and when these regions are subject to different aerodynamic, inertial and system loads happening during different in-flight and in-ground maneuvers \cite{stress, stress2,stress3}.
Accordingly, input data $\bf X$ consists of $n=26$ features (external loads applied to a specific region of the aircraft, and different maneuver specs \cite{stress3}). 

\section{Validation pipeline}
\label{sec:pipeline}
\begin{figure}[htb]
\begin{centering}
\includegraphics[width=0.9\textwidth]{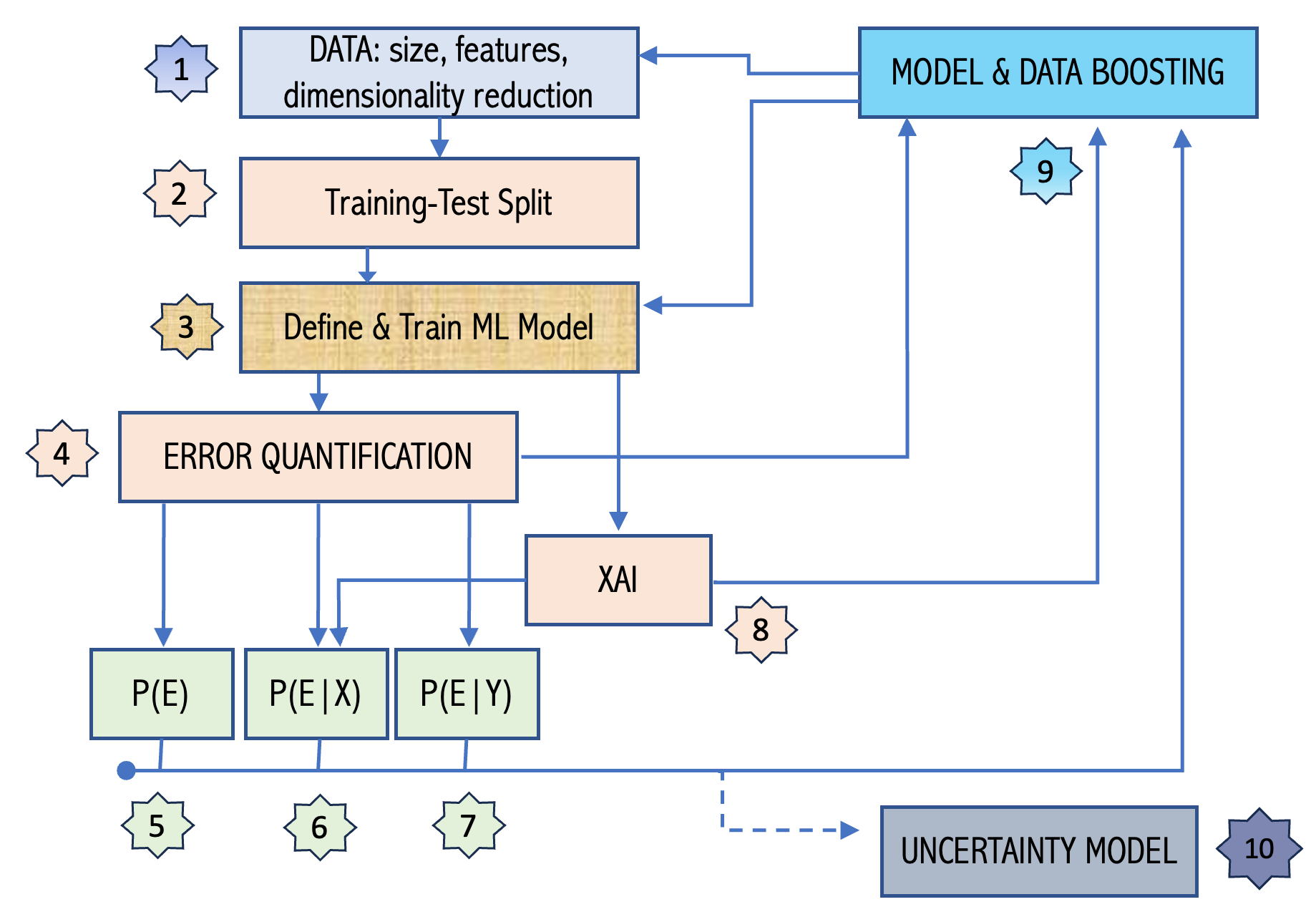}
\par\end{centering}
\caption{Sketch of the industrially-tailored validation pipeline of a Machine Learning model we propose in this paper. Each of the boxes is an important part of the whole process and is detailed as a specific subsection in the text. Observe that the pipeline is a directed graph with several cycles, that illustrate the different iterative refinements of the whole process.}
\label{fig:pipeline}
\end{figure}

Whereas one can assess the quality of the trained ANN by something as simple as to compute the error function (Eq.\ref{eq:Error_MSE}) in the test set, or to assess the correlation in a scatter plot between $\bf{\tilde{Y}}$ and $\bf Y$ in the test set, usually the validation of industrial problems requires substantially more analysis, to gain understanding and trustfulness of the machine learning solution. In Fig.\ref{fig:pipeline} we sketch our proposed validation pipeline, that includes a sophisticated statistical validation part, designed to be implemented in industrial problems requiring a supervised learning approach. In what follows, each subsection will explain the specifics of each box within this pipeline, and we will provide illustrative examples of it within the industrial problem of reference.

\subsection{Data}
\label{sec:data}
This corresponds to box $\#1$ in Fig.\ref{fig:pipeline}.

\medskip
\noindent {\bf Data uncertainty and data augmentation --} The first step of the pipeline deals with building the dataset $\text{Full}=\{({\bf X}(i),{\bf Y}(i))\}_{i=1}^N$. Ideally, the full dataset should appropriately cover the whole feature space, or more specifically cover the subspace of interest (see Section \ref{sec:boosting} for a discussion). A standard technique to guarantee that such subspace is adequately covered by the $N$ datapoints --specially useful when having large values of $N$ is costly-- deals with using high space-filling sampling methods such as Latin Hypercube Sampling \cite{LHS0, LHS} to select the combination of feature values of these $N$ datapoints, although other choices can be used depending on the problem desiderata and engineering constraints.\\
For each point, observe that $\bf Y$ is a `ground true' value, i.e. the dataset is made by duples of input/outputs which the ANN will assume they represent the true function it aims to find. In industrial problems, such dataset is usually built based on a variety of methods. For instance, they could be the result of extensive numerical simulations, and/or wind tunnel experiments, etc (when different sources of data are used, these are called multifidelity datasets, as a priori the quality of data can vary). The standard, basic approach in Machine Learning is to assume that the dataset is error-free, i.e. the duple $(x,y)=(0.12,0.7)$ states that for an entry value of 0.12, the output is deterministically equal to 0.7. Now, one could also attach to each of the data points a given intrinsic uncertainty, which could explicitly state the fact that these are the result of experiments or other procedures that generate measurement, numerical or other types of errors (specially useful in multifidelity datasets). Accordingly the same duple can read $(x,y)=(0.12 \ \text{CI95} \ [0.09,0.15], 0.7 \ \text{CI95} \ [0.69,0.71])$. This means that the entry data cannot be guaranteed to be equal to $0.12$ (e.g. due to some experimental measurement error): it is indeed a random variable whose distribution's 2.5 percentile is at 0.09 and the 97.5's percentile is at 0.15, i.e. with probability 0.95 the true entry data is somewhere between 0.09 and 0.15. Similarly, the output data might have some measurement error, or some numerical error, such that the true output lies, with probability 0.95, between 0.69 and 0.71. 
If one decides to incorporate such kind of uncertainties in the dataset, one proceeds to `sample' uncertainty-free replica tuples: we assume that $x$ and $y$ are independent random variables with the prescribed $95\%$ confidence intervals, independently extract $q$ samples from these random variables, and combine these to build $q^2$ new, error-free data duples\footnote{The assumption of independence is parsimonious. If $x$ and $y$ are explicitly correlated via a certain co-variance expression, this can be used in the sampling process}. This is indeed a classical method of data augmentation (see section \ref{sec:boosting}).

\medskip
\noindent {\bf How many features? --} We also need to consider which are the features that will build up the feature vector ${\bf X}=(x_1,x_2,x_3,\dots,x_n)$. In theory, all physically or industrially-relevant features must be taken into account, so the larger $n$, the better. However, in order for the ANN to have a good performance throughout all combinations of the input variables, it needs to be trained with sufficient data that, in particular, accurately sample the whole feature space. It is thus easy to see that the number of datapoints $N$ to accurately sample $n$ features scales exponentially with $n$: the hypervolume of an $n$-dimensional cell of length $L$ is $L^n$, and thus assuming that we need at least one datapoint for cell, $N\sim O(L^n)$. Accordingly, we need to strike a balance: $n$ needs to be large enough so that most relevant aspects that inform or affect the output $\bf Y$ are included, but small enough so that the dataset of size $N$ is sufficiently sampling the feature space of dimension $n$. There are just two ways: either we increase $N$, or we reduce $n$. Increasing $N$ is in most of the cases very expensive, as it requires to perform more experiments, or run additional and costly simulations (one can proceed to perform data augmentation as an alternative, although this choice does not necessarily help sampling better the feature space). Reducing $n$ can be done via linear dimensionality reduction (see below), or eventually by leveraging XAI methods (see sections \ref{sec:xai} and \ref{sec:boosting}).

\begin{figure}[htb]
\begin{centering}
\includegraphics[width=0.5\textwidth]{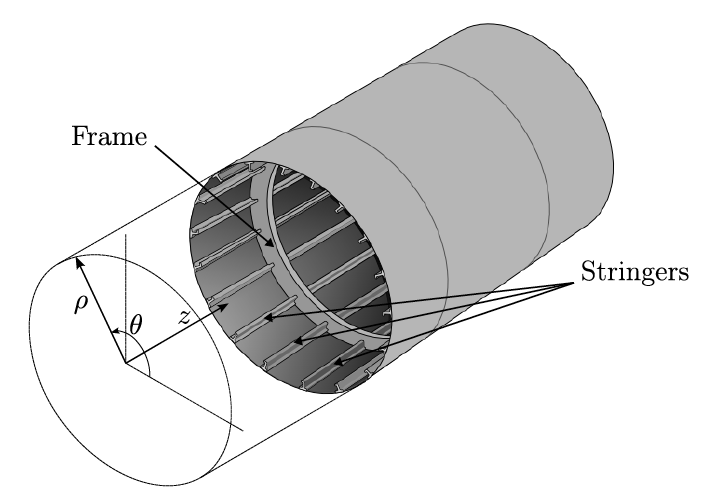}
\par\end{centering}
\caption{Illustration of frames and stringers in an aircraft's fuselage section, along with a cylindrical coordinate system. Image edited by the authors based on \cite{imagen_superstringer}, under Creative Commons Attribution 3.0 Unported.}
\label{fig:superstringer}
\end{figure}

\medskip
\noindent {\bf Encoding --} By construction, some of these features can be numerical, while some other can be categorical. An important step is thus to assess whether we need to encode categorical variables as numerical ones. This step is often required, for instance, if one aims to perform some dimensionality reduction of the ambient space. Additionally, the entries of the ANN are usually numeric and therefore the encoding might be required when training the model (section \ref{sec:model}). Observe that one can one-hot encode any type of categorical variable, but this step is quite inefficient for variables that have a large number of categories, as it enlarges the ambient space dimension very quickly.
Accordingly, often purely categorical variables are not included in the subsequent analysis. There are, however, categorical variables that one can encode using more efficient strategies. Consider for instance the problem of stress engineering. Some of the relevant entry data include different types of {\it loads}, which are measure of forces and therefore numerical. Additionally, the specific region of the aircraft could be parametrized, for instance as a combination of a {\it Frame} location and a {\it Stringer} location. These are usually categorical, e.g. Frame86-Stringer12 (see Fig.\ref{fig:superstringer} for an illustration). However, these categories have a geometrically-induced ordering: frames are ordered along the fuselage, whereas stringers have a polar symmetry, and thus the duple frame-stringer has a natural cylindrical embedding. Therefore, one can easily encode frame using an ordinal variable (the $z$ axis in the cylindrical embedding). Stringers can be efficiently encoded using the polar angle $\theta$: if we have $q$ stringers, then a certain stringer location Stringer12 maps into a two-dimensional variable 
$$\text{Stringer12} \to (\cos(\theta_{12}), \sin(\theta_{12})),$$ where $\theta_x=2\pi x/q.$

\medskip
\noindent {\bf Data normalization --} If the range of each of the features in the data is very different, then any subsequent method applied to either find patterns in the data or leverage those patterns for prediction will have a hard time to adequately resolve the contribution of features with very different ranges. In order to solve this issue, often data are {\it normalized}. This basically means that each feature is normalized individually over all the data, such that features have comparable ranges (one can do min/max normalization, data standardization, or other choices). An often important remark is that, in order to avoid data leakage, the normalization process (e.g. calculating the min, max, mean, variance, etc) of each feature  needs to be performed in the {\it training} set, and the resulting quantities need to be used to then normalize the {\it test} set, rather than computing these quantities over the whole dataset.

\medskip
\noindent {\bf PCA of input data --} In order to try and reduce the effective number of features $n$ is also usually helpful to linearly reduce the dimensionality of the ambient space by performing a Principal Component Analysis (PCA) \cite{ML} on the data matrix $X \in \mathbb{R}^{N_{\text{training}}\times n},$ i.e. a matrix with as many rows as elements on the training set, and as many columns as (numerically encoded) features. PCA performs an affine transformation of the coordinate system so as to find an orthonormal base whose unit vectors are ordered according to the data variance that is spanned through that vector's direction. If the features show some linear correlations, then data actually live in a smaller subspace than the original feature space: the PCA procedure finds the adequate linear combination of original features that form each new, `combined features' called principal components, and we typically need less than the original $n$ features to describe most of the variance. Depending on the industrial problem under analysis, the required number of principal components (and thus the effective reduction) can vary, but most often than not, the physics imposes different types of forces to be correlated, so that PCA is usually helpful.\\
As it happens with data normalization, PCA needs to be performed in the training set, and then the test set data need to be projected into PCA space of the training set (rather than performing PCA on the test set or performing PCA on the whole dataset).

\subsection{Train-Test Split: when is this done adequately?}
\label{sec:split}
This corresponds to box $\#2$ in Fig.\ref{fig:pipeline}.

\medskip
\noindent The most basic way to train the ANN and test its performance is by splitting the full dataset into a training and a test set (usually a 80/20 split), optimizing the parameters of the ANN based on using only data from the training set, and testing the performance of the optimized ANN in unseen data by computing the error (e.g. Eq.\ref{eq:Error_MSE}) of the ANN in the test set. This is the so-called Holdout (80/20) rule.
There are other, more sophisticated methods, such as k-fold crossvalidation, often used when the full dataset is not too large \cite{ML}. Here we focus on the Holdout case, which only entails a basic split. Now, this split of the full data between training and test ($\text{Full}=\text{Train}\cup\text{Test}$) needs to be done `adequately'. This is necessary to make sure that the model can be trained as much as possible, and that the error found in the test set is accurately capturing the model's performance and generalization capability.  What could go wrong, otherwise? Here we outline a few examples:
\begin{itemize}
    \item if the test set only includes data from a small region of the feature set, testing will only analyse the performance of the ANN in that small region. If that region results to be properly represented in the training set, the test error would be giving the wrong idea that the ANN generalized very well, where the reality is that the ANN works well in that small region of the input space.
    \item if the test set includes data which are extremely similar to data in the training set, the testing error will be closer to training error, i.e. we wouldn't be properly evaluating the capacity of the ANN to generalize (i.e. to correctly predict the output associated to unseen entry data). The ANN could be massively overfitting and misleadingly giving a low test error.
    \item if the training set does not include data that adequately sample the feature space, the ANN wouldn't be able to learn poorly sampled regions, hence anticipating very poor generalization for these regions.
\end{itemize}

We define a TT split as {\it adequate} if (i) the frequency of having the same datapoint in both the train and test sets is vanishingly small, and (ii) the vector distributions of both sets are similar, i.e.
\begin{equation}
    \label{eq:TTcondition}
    P([x_1,x_2,\dots,x_n]^{\text{Train}}) \,{\buildrel d \over =}\, P([x_1,x_2,\dots,x_n]^{\text{Test}})
\end{equation}

Popular wisdom in Machine Learning assumes that a random partition --i.e.  randomly assigning with probability $p$ elements from the full dataset into the training set, the remaining being left in the test set-- is a method that yields adequate TT splits (concretely, $100p\%/100(1-p)\%$ TT splits).
This wisdom is loosely based in the fact that uniform random sampling makes, when $N\gg n$ and in the limit of large $N$, two datasets where Eq.\ref{eq:TTcondition} is asymptotically verified. However, the convergence is {\it slow} when $n$ is large, and thus, in practice a random TT split is not necessarily always working as expected. Another splitting technique involves first setting the TT split percentages, and then try and sample from the full dataset the training and test sets following Lating Hypercube Sampling. 
In any case, before blindly accepting any given TT split, one needs to make sure that such split is ``adequate''. Here we provide a few preliminary comments and desiderata on this task, and then introduce a method to assess such split.

\medskip
\noindent Because neural networks are good interpolators but a priori poor extrapolators, test cases are expected to be somewhat ``inside'' the space covered by training, i.e., in the interpolation regime. Mathematically, interpolating regime amounts to convex hull membership \cite{CHMP1}. Let $\texttt{Hull}(\text{Train}_X)$ be the convex hull of the set of points formed by all the input data $\bf X$ of the training set, and  $\text{Test}_X$ the set of points formed by all the input data of the test set
(i.e. for each duple $({\bf X}, {\bf Y})$, we select only the point ${\bf X}$; observe also that ${\bf X} \in \mathbb{R}^n$, that is, if needed, encoding of categorical variables has been made). Then we have the following definition. 

\medskip
\noindent {\bf Definition 1} (Interpolating Regime). The ANN is said to be in interpolating regime if, $\forall {\bf X} \in \text{Test}_X: {\bf X} \in \texttt{Hull}(\text{Train}_X)$. 

\begin{figure}[htb]
\begin{centering}
\includegraphics[width=0.9\textwidth]{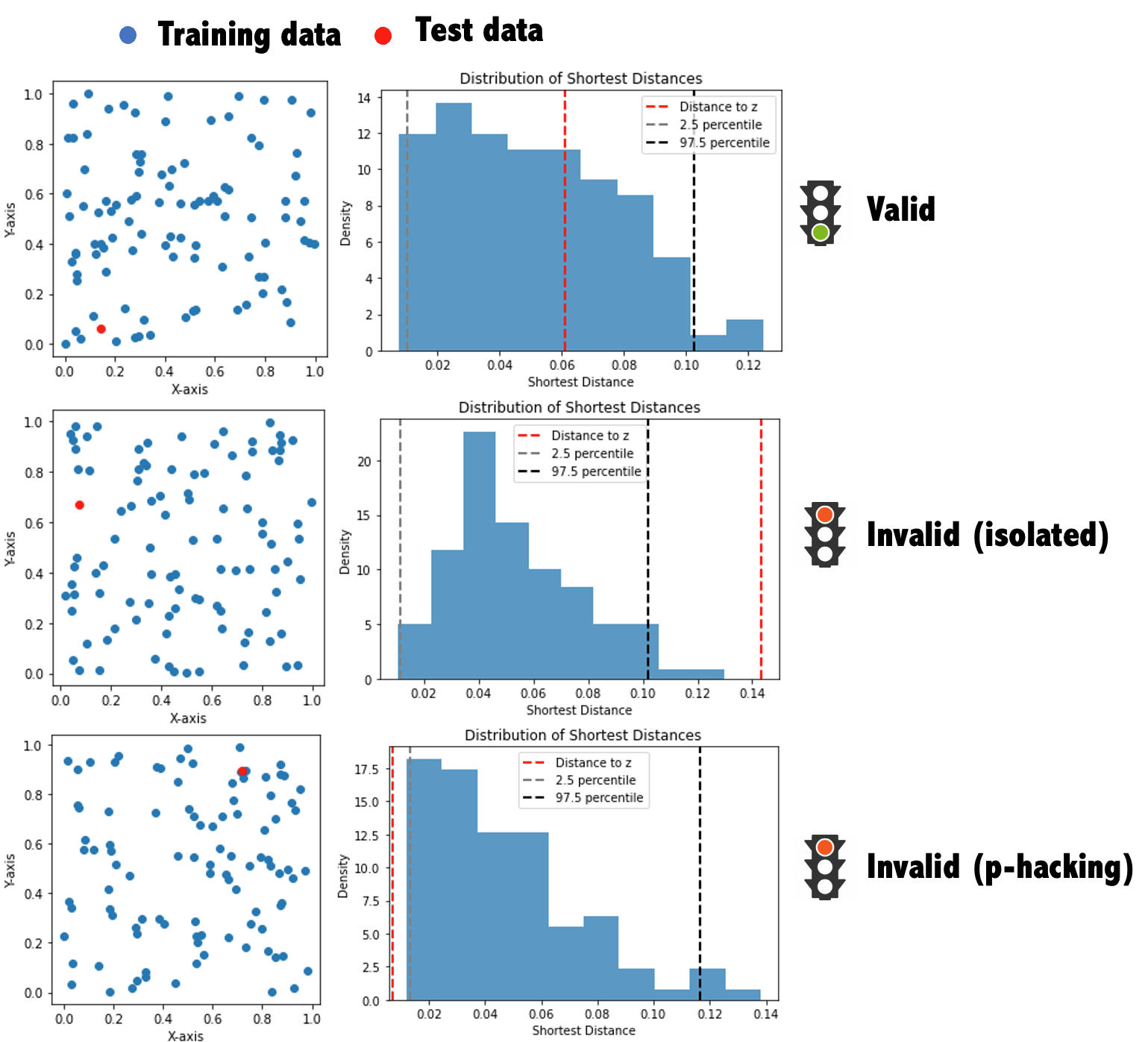}
\par\end{centering}
\caption{Illustration of the proximity method, where within a given data voxel (here a 2-dimensional box) populated by 100 training data, we compute the shortest distance between a given test datum (red dots) and the voxel's training set (blue dots). This distance is then compared to the distribution of shortest distances found by computing the shortest distance between any pair of training set data within the voxel. The test datum is deemed `valid' is its shortest distance to the voxel's training set is within the 95\% confidence interval, i.e. between the 2.5 and 97.5 percentiles (first row). Otherwise the test datum is deemed 'invalid': those test data whose distance is above the 97.5 percentile (second row) are classified as within an isolated region inside the voxel (and therefore will lead to poor interpolation), whereas if the distance is below the 2.5 percentile (third row), the test data is classified as performing p-hacking, i.e. it is too close to the training set.}
\label{fig:proximity}
\end{figure}

\medskip
\noindent Requiring the ANN to be in the interpolating regime seems to be a sensible a priori to be a necessary condition for the ANN to adequately predict, and thus could be seen as a first test that can be performed to evaluate the quality of the TT split. However, observe that in high dimension $n$, computing $\texttt{Hull}(\text{Train}_X)$ is computationally very expensive. The so-called convex hull membership problem (which does not require to explicitly compute the convex hull, only to decide Definition 1) is computationally cheaper, but still costly for large $N$, as it is the case in industrial problems, and thus this does not seem a viable way forward. In fact, the computational cost is not the only problem with this approach. Observe first that $\texttt{Hull}(\text{Train}_X)$ could have isolated regions or simply data holes --i.e. regions inside the convex hull that are disproportionally less densely filled by training data--. If a test data point falls within one such isolated region, even if this is inside $\texttt{Hull}(\text{Train}_X)$, it is not obvious that the ANN will interpolate adequately (in other words, membership is not a sufficient condition). Second, as $n$ increases, the curse of dimensionality quickly sets in. Interpolation in large dimensionality is indeed very hard, as $\texttt{Hull}(\text{Train}_X)$ occupies an exponentially shrinking volume as compared to the volume of the $n$-dimensional hypercube spanned by the ranges of each of the input variables. In fact, recent works \cite{lecun1} even suggest that ANNs are consistently working in the extrapolation regime, casting doubts on the actual relevance of the ANN being in the interpolation regime for being able to adequately generalize. All these issues lead us the necessity to (i) build a computationally efficient protocol, (ii) which is able to evaluate whether test data are actually reasonably close to the training data so that the interpolative power of the ANN can be leveraged, even if that does not necessarily entail a convex hull membership. 

\medskip
\noindent {\bf Voxel tesselation and proximity method (VTPM)--}
Here we propose to make use of a simple and efficient protocol that we call the {\it voxel tesselation and proximity method} (see Fig.\ref{fig:proximity} for illustration). In a nutshell, the algorithmic steps of the method are as follows.

\begin{figure}[htb]
\begin{centering}
\includegraphics[width=0.8\textwidth]{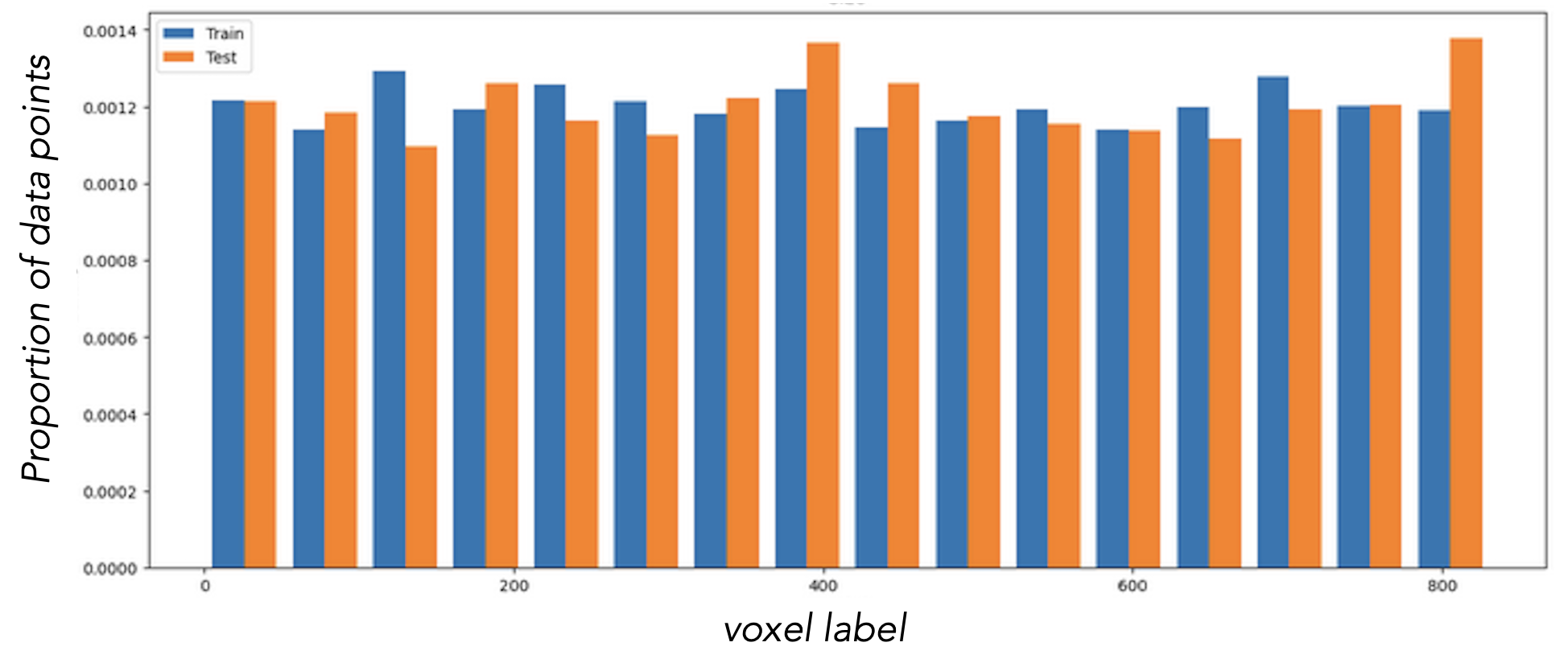}
\par\end{centering}
\caption{Number of datapoints of the training (blue) and test (orange) sets inside each voxel. There are about 800 voxels. For better visualization, voxels have been artificially binned.}
\label{fig:voxel}
\end{figure}

\begin{enumerate}
\item The first step is to build the voxel tesselation of the training set. Voxelization is a way of tesselating the training set according to the categorical variables only. Accordingly, a given voxel is a specific subset of the training set composed by all the training data for which the categorical variables take particular values. Hence the total number of voxels is equal to the product of the number of categories of all categorical variables.
\item Once this voxelization has been made, all test set points are assigned to one of these voxels, according to the values of their categorical variables. If the categorical coordinates of a given test point $z$ are not represented in the training set, an additional voxel --called the {\it residual voxel}-- is created and $z$ is assigned to this extra voxel. Data in the residual voxel will be used later on for data-augmenting the training set (see section \ref{sec:boosting}), so as to create and populate the necessary additional voxels.
\item Accordingly, the voxel distribution for the training and test sets is computed, by expressing the percentage of points in each voxel. If there are more than $5\%$ of points of the test set in the residual voxel, we conclude that the TT split is not adequate. Otherwise, a two-sample $\chi_2$ test is used to assess the quality of the test set covering the training set. Ideally, the $p$-value should be larger than 0.05, i.e. we cannot reject the null hypothesis of equal distributions. If the null hypothesis is rejected, we will explore which voxels have a disproportionally small number of training points (as compared to test points). These voxels will be data-augmented (see section \ref{sec:boosting}). \\
Now, for each test set point $z$:
\item we compute the shortest distance $d_z$ between the test data under analysis and the voxel's training data subset (effectively, this is just the minimum over the set of shortest distances between $z$ and any training data in the voxel).
\item We then compute the shortest distance between all pairs of the voxel's training data. We extract the 2.5 and 97.5 percentiles of the resulting distribution, labelled P2.5 and P97.5 respectively
\item If $P2.5<d_z<P97.5$, $z$ is classified as a `valid' test datum. If $P97.5<d_z$, $z$ is classified as `invalid/isolated', whereas if
    if $P2.5>d_z$, $z$ is classified as `invalid/p-hacking'.
\item Finally, this analysis is performed  $\forall z \in \text{Test}$. The TT split is considered adequate if the percentage of `valid test points' is larger or equal to $95\%$.
\end{enumerate}
For illustration, Fig.\ref{fig:voxel} illustrates the partition of training and test sets into voxels, showing that such partition is roughly similar for the train and test set, as expected. In this example, the percentage of valid test set points is larger than $95\%$, hence the split is deemed adequate.

\subsection{Definition and training of the ML model}
\label{sec:model}
This corresponds to box $\#3$ in Fig.\ref{fig:pipeline}.

\medskip 
\noindent {\bf Choosing the architecture and the loss --} It is well known \cite{DL} that different algorithms and learning architectures are better designed for certain types of supervised learning tasks. In the context of artificial neural networks, it is well known that (deep) feed-forward networks are suitable architectures for generic data prediction. Data that has a {\it spatial} structure (e.g. image classification) often benefit from a network architecture with a convolutional layer \cite{DL}, resulting in a convnet solution, whereas data that have a {\it temporal} dependence (e.g. time series or text) often benefit of a network architecture with cycles, resulting in recurrent networks (e.g. reservoir computing, echo state networks). If the data itself has a {\it relational} structure, then the preferred architecture is that of a graph neural network.\\
For illustration, the industrial case under analysis aims to predict reserve factors of $m=6$ different failure modes at specific regions of the fuselage and for different combinations of loads. If we treat the data as a raw set of features, we could use a (deep) feed-forward network, e.g. a multilayer perceptron. However, observe that the categorical variables associated to the region of the aircraft (combination of frame and stringer) encode spatial information. We can therefore use some specific architecture that leverages spatial information, such as some kind of convolutional layer.\\
As for the loss function, we initially use MSE as the error term, and impose a $L_2$ regularization. Later on (section \ref{sec:boosting}) we will discuss how to improve the loss function by refining both the error term and by adding more regularization terms, such as forcing the $m=6$ predicted reserve factors to preserve a certain correlation structure that emerges from the physics of the problem (i.e. failure modes are not independent).

\begin{figure}[htb]
\begin{centering}
\includegraphics[width=0.5\textwidth]{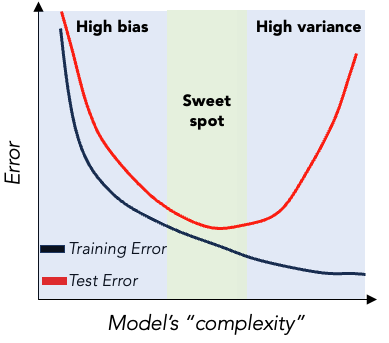}
\par\end{centering}
\caption{Sketch of the surrogate model's errors (e.g. mean square error as per Eq.\ref{eq:Error_MSE}) applied to the training set and test set, as a function of the surrogate model's `complexity', i.e. loosely speaking the number of tunable hyperparameters, showcasing the bias-variance trade-off paradigm: when a model is not complex enough, it cannot learn the patterns in the data, and thus both training and test errors are high (high bias region). When a model is unnecessarily complex (severely overparametrized), it will overfit and thus will have a very low training error but a large test error (as the model learnt not only the patterns in the data but also the random irregularities in the training data, which do not systematically appear in the test data). This is the high variance regime. A good model is the one that strikes a balance.}
\label{fig:bias_variance}
\end{figure}

\medskip 
\noindent {\bf How complex? Bias-variance tradeoff --} Once the architecture and the loss have been decided, we need to decide how complex the model ought to be. For instance, in a Feed-Forward network, one should decide: how many layers? How many neurons per layer? In other words, how many trainable parameters? How complex? It is easy to see that if the machine learning model is `too simple' (i.e. too few parameters), it could be the case that the model is just incapable of learning the patterns hidden in the training set. The model is simply too rough  --it simply underfits the training data-- and, even after being optimized, the training error (e.g. Eq.\ref{eq:Error_MSE}) is very large. This type of error is called a bias error. In this circumstance, the test error --e.g. Eq.\ref{eq:Error_MSE} performed in the test set-- will necessarily also be very large, if only because it is making the same systematic mistakes than in the training set. In this situation, we say that the model has `high bias', see Fig.\ref{fig:bias_variance} for an illustration. 
At the other extreme, it is easy to see if that the machine learning model is `too complex' for the learning task (i.e. severely overparametrized), it very well can be that the model not only is capable of learning the patterns hidden in the data, but its `extra' degrees of freedom are then used to also learn the noise (i.e. irregularities which are not part of the pattern) in the training data. The bias error is clearly very low in the training set, but the model is overfitting the training data. Accordingly, when confronted to the unseen data in the test set, despite the fact that the model correctly learnt the pattern, it also learnt the random irregularities of the training data, and will force these in the test set as well, irremediably causing the test error to be large. This kind of error is called the {\it variance error}. In this situation, we say that the model has `high variance' (see Fig.\ref{fig:bias_variance}, classically depicted as a large gap between training and test errors).\\
The correct model strikes a balance between high bias and high variance: we aim for a small training error but also a small test error. 
Finding this balance --the so-called bias-variance tradeoff-- is a difficult problem, a bit of an art really, and the process of optimizing it is called hypertuning \cite{DL}, which we won't discuss here. What is important to have in mind is that there is indeed a way to systematically evaluate whether our model choice is showing high bias, high variance, or if it is striking such bias-variance tradeoff, and to understand whether increasing the training set size could have an effect in this tradeoff, by assessing the so-called {\it learning curves}. This is one of the techniques we will explore for refining and boosting the model This part of the pipeline is discussed in section \ref{sec:boosting}.

\medskip 
\noindent {\bf For how long? --} In the training of neural networks, a natural way to parametrize the number of iterations of the optimization scheme is the number of epochs. If we draw a graph of the training and test errors as a function of the number of epochs of the optimization procedure, one can end up finding something similar to Fig.\ref{fig:bias_variance}, where the X axis encodes the number of epochs instead of the model's complexity. The idea is that if one optimizes for too many epochs, the model can start to overfit to the training set. This will produce systematically smaller training errors, but when overfitting kicks in, the test error will start to increase. It is not clear a priori for how long a given surrogate model needs to be trained, but drawing error-vs-epochs curves --and finding the number of epochs for which the test error has a flat minimum-- is a technique to set this hyperparameter.

\subsection{Point-wise global error quantification}
\label{sec:error}
This corresponds to box $\#4$ in Fig.\ref{fig:pipeline}.

\medskip
\noindent As mentioned previously, the simplest approach to evaluate the performance of a trained ANN is just to measure some pointwise error metrics on the test set. The classical ones are the Mean Absolute Error (MAE, Eq.\ref{eq:Error_MAE}) and the Mean Square Error (MSE, Eq.\ref{eq:Error_MSE}). 
Having similar error metrics in the training and test sets is a good indication that the model doesn't have high variance. If such error is large, then the model is probably not complex enough, or the regularization term is too strong (see section \ref{sec:boosting} for details on decision making with respect to model refining).
Besides being plain and easily interpretable scalar metrics of the model's performance, MAE and MSE are also used when building the so-called learning curves (see again section \ref{sec:boosting}) that provide further information on the quality of the model and the margin of improvement when larger datasets are considered.\\
Now, while both provide a net magnitude of the mismatch between predictions and ground true values in the test set,  
by construction they do not provide information of whether mismatches are due to over- or under-predicting (i.e. larger or smaller than ground truths). In fact, in some industrial problems the specifics of the mismatch can be very important. For instance, when aiming to predict reserve factors of failure modes, having a mismatch due to the fact that the ANN is predicting systematically smaller reserve factors (the ANN is overestimating the likelihood of a failure mode, i.e. it is being over-conservative) is definitely `better' than if the mismatch is caused by predicting systematically larger reserve factors (i.e., the ANN is underestimating the likelihood of a failure mode). This necessary distinction leads us to use the concept of Residual Error, defined as the (signed) discrepancy between the actual, ground true value and the value predicted by the model. The {\bf Residual Error} or simply Residue of a given point $i$ is 
\begin{equation}
    {\bf e}(i)={\bf Y}(i) - {\bf \tilde{Y}}(i).
    \label{eq:residue}
\end{equation}
The residue of the $j$-th component of the output variable is just $e(i)=y_j(i)-\tilde{y}_j(i)$. It is easy to see that the MSE is essentially built by averaging the squared residue --i.e., MSE is the first moment of the distribution $P(e^2)$-- whereas for MAE the average is over the absolute value of the residue. At this point it is important to highlight that while a null MSE is indeed pointing to a perfect performance of the model, this is not true for the average of the residue over samples, as residues can cancel out. In fact, a null average of the residues is simply informing that the mismatches are symmetrical, i.e. there are as many underestimations as there are overestimations. Accordingly, the residue is not really helpful to build a pointwise quantification of the global error. However it will be important in what follows, when we reconstruct residue distributions. 

\begin{figure}[htb]
\begin{centering}
\includegraphics[width=0.5\textwidth]{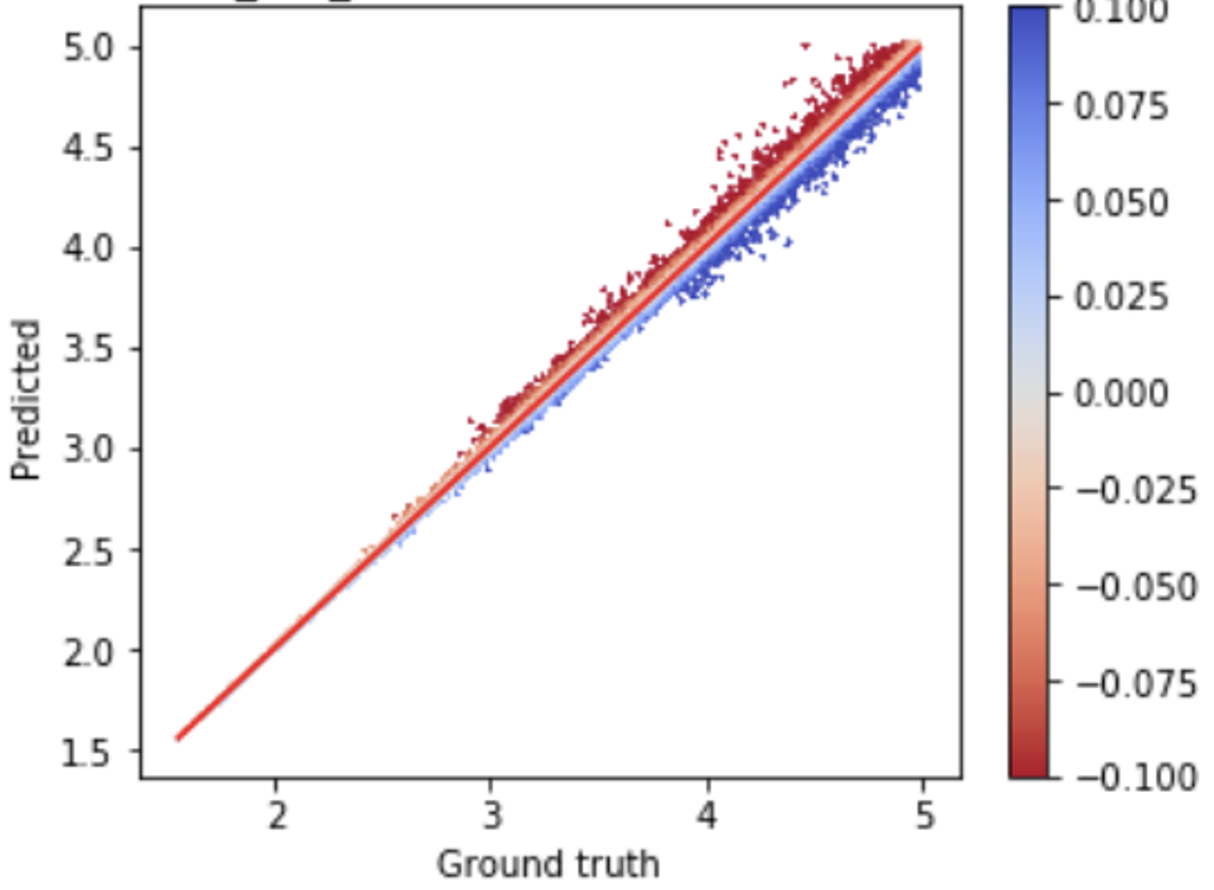}
\par\end{centering}
\caption{Scatterplot between ground true and predicted values of the reserve factor associated to a specific failure mode, for all data configurations in the test set (test set size is roughly about $10^6$ datapoints). The $R^2=0.999$, highlighting that the model performs very well. Dots are color-coded according to whether there residues are positive (conservative model) or negative (underestimates risk, dangerous). 
This scatterplot also shows a certain degree of heteroscedasticity: the residue seem to increase for larger values of the reserve factor.}
\label{fig:scatter}
\end{figure}

\medskip
\noindent Finally, another important scalar metric assessing globally the performance of the ANN is the $R^2$ of the scatterplot $y_j$ vs $\tilde{y}_j$, for each component of the output variables $j=1,2,\dots,m$. $R^2$ provides an indication of the goodness of fit of a set of predictions to the ground true values. In the statistical literature, this measure is called the coefficient of determination. For illustration, Fig.\ref{fig:scatter} depicts a scatter plot of $y$ vs $\tilde{y}$ where $y$ is the reserve factor of a specific failure mode, for all the points in the test set. The $R^2=0.999$, suggesting that the model performance is very good. Colors highlight whether the residue is positive (overestimates risk, e.g. conservative model) or negative (underestimates risk, undesirable).

\subsection{Marginalized residue distribution}
\label{sec:error_marginal}
This corresponds to box $\#5$ in Fig.\ref{fig:pipeline}.

\medskip
\noindent The reason why pointwise error metrics --like MSE or MAE-- are often used is not only for simplicity: they often have the implicit assumption that individual errors will not severely deviate from the average case: the average case is an informative statistic. Now, is this always the case? In practice: unfortunately no. 
In terms of the residue, the implicit assumption is that the residue distribution is reasonably close to a Gaussian distribution. But in industrial applications, it is easy to conceive data configurations where the model performs very well, and other configurations where the prediction is very bad. This heterogeneity is, to some extent, related to some of the problems discussed in section \ref{sec:split} about the interpolating capacity of neural networks, and the difficulty of accurately sampling the feature space with training data when the number of features is somewhat large. It is quite common, in problems requiring the feature number $n$ to be in the order of the dozens, to unavoidably have regions of the feature space better sampled in the training set than others. Another reason why the performance of the ANN can be heterogeneous is because there might be specific (input) data configurations which are associated to strong physical gradients: regions of the feature space where the physics is more difficult to capture --e.g. it is more nonlinear-- and the interpolative nature of the neural network outputs is not enough to provide an accurate prediction. Overall, this can result in heterogeneous errors, and thus, the notion of an average error can break down. This is the first motivation to go beyond pointwise error metrics, and reconstruct error distributions instead.

\begin{figure}[htb]
\begin{centering}
\includegraphics[width=0.9\textwidth]{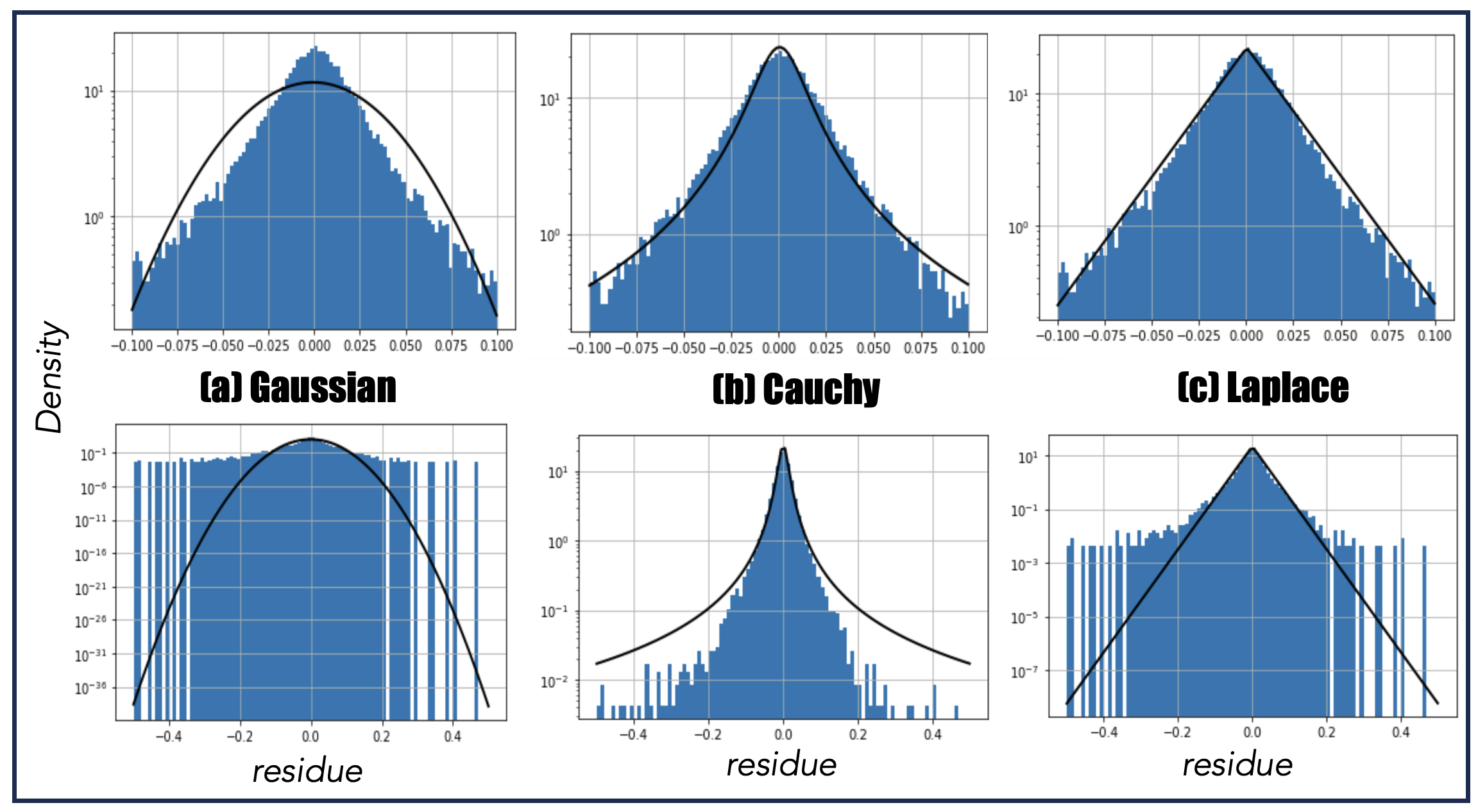}
\par\end{centering}
\caption{Log-linear plots of the probability density distribution of error residuals in the test set. The panels in the top row focus on the range of residues $[-0.1,0.1]$, whereas the panels of the bottom row display the whole residue range, highlighting the appearance of extreme events. The same data are plotted in all panels. The three columns also display (solid black line) the best fits to different distributions: (a) a Gaussian distribution, (b) a Cauchy distribution, and (c) a Laplace distribution.}
\label{fig:P(E)}
\end{figure}

\medskip
\noindent The second reason why this endeavor is not only sensible but necessary is because from the reconstructed error distributions one can perform outlier detection. In a nutshell, an outlier is a datapoint (i) whose presence is considered unlikely, or (ii) whose position is considered far from expected. Classical outlier detection methods are thus based on having access to the marginal distribution of the data (here, the residue distribution), so that the probability $p_{>x}$ of observing a residue larger or equal to some observation $e_x$ is simply $p_{>x} = \int_{e_x}^{\infty} P(e)de$. If, say, we have $N$ observations and $p_{>x}N < 1$, then residues larger or equal to $e_x$ shouldn't be found... hence labelling all the residues larger or equal to $e_x$ as outliers. This is just a simple heuristic, but several outlier detection schemes are based on similar arguments.\\
The third reason why reconstructing error distributions is a good idea is because from them one can extract uncertainty measures, that ultimately will inform how much to trust a model's prediction by associating a confidence or a prediction interval (see section \ref{sec:uncertainty} for details on how to build and validate a complete uncertainty model). Computing these intervals accurately classically requires fitting the reconstructed error distribution to a list of parametric distributions, with the Gaussian distribution as the first in the list. In the event that parametric distributions are not properly representing our data, then the method of choice is applying some nonparametric procedures (e.g. nonparametric bootstrapping) to estimate the intervals of interest.

\begin{figure}[htb]
\begin{centering}
\includegraphics[width=0.55\textwidth]{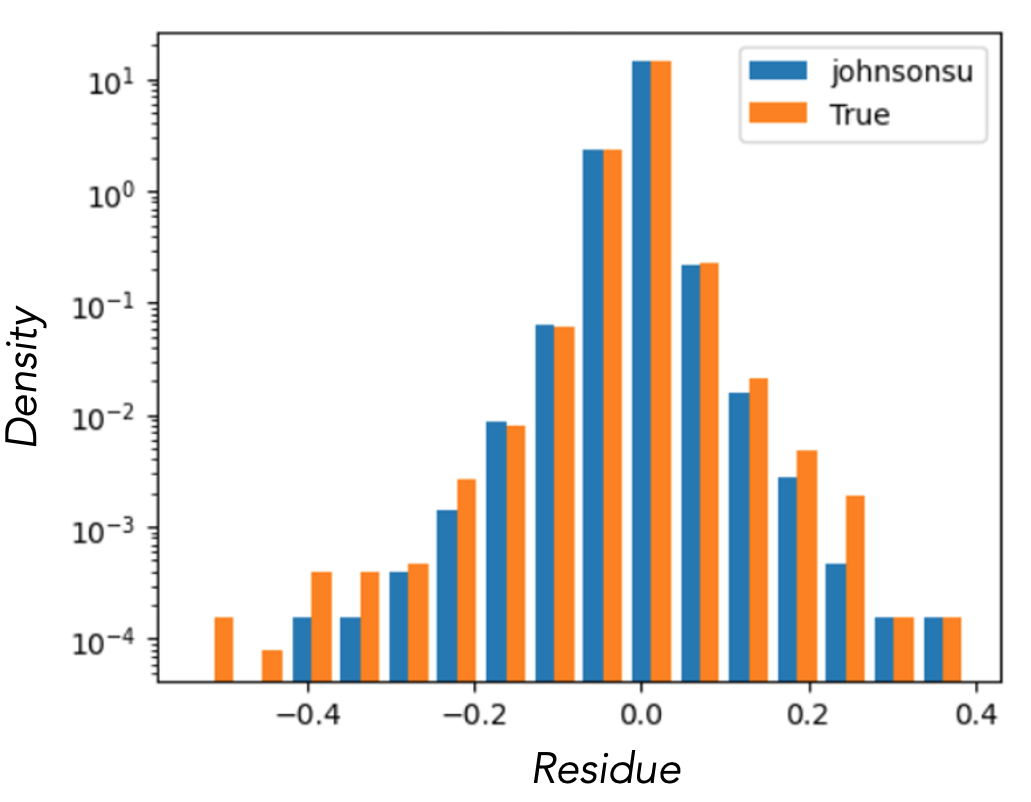}
\par\end{centering}
\caption{Orange bars are a log-linear barplot of the probability density distribution of error residuals in the test set (from the same data as in Fig.\ref{fig:P(E)} but with a different binning, to allow a better visualization). Blue bars are the result of fitting a Johnson's SU distribution to the data, generating a number of points equal to the test set size from this distribution, and reconstructing the pdf with the same bins as the empirical one.}
\label{fig:johnson}
\end{figure}

\medskip
\noindent As an illustration, in Fig.\ref{fig:P(E)} we plot the empirical residue probability distribution function obtained, for residues of the reserve factor of a certain failure mode in the test set, along with the best fit to three parametric distributions: a Gaussian distribution (panels a), a Cauchy distribution (panels b) and a Laplace distribution (panels c). After performing a Kolmogorov-Smirnov test and an Anderson-Darling test\footnote{The Anderson-Darling test gives more weight in the calculation of the test statistic to the distribution tails, as compared to the Kolmogorov-Smirnov test} to each parametric distribution, we conclude that none of the three offer good fits ($p$-value$<0.001$). If we focus on the central residue range $e\in[-0.01,0.01]$ where most of the residues lie, we clearly see that the residues are not Gaussian, but the Laplace distribution visually fits the data very well. However, if we zoom out and consider the whole residue range (bottom row), we see that both the Gaussian and also the Laplace fits underestimate the appearance of data with very large residual error (either positive or negative), i.e. these distributions cannot explain the onset of extreme events. These extreme events are important to capture: while rare, some correspond to configurations where the ANN dangerously underestimates the failure risk (negative extreme events) and these need to be considered.  
For comparison, we also show the fit to a Cauchy distribution, that in this case massively overestimates the presence of extreme events. If we were to build an uncertainty model based on the Cauchy distribution, the uncertainty associated to each prediction would be unnecessarily large.\\
Anecdotically, in this case we also found an (exotic) parametric distribution whose fit indeed passed a Kolmogorov-Smirnov test: this is the Johnson's SU distribution \cite{johnson}, a four-parameter family $\text{JohnsonSU}(\gamma, \delta, \xi, \lambda)$ successfully used to model asset returns for portfolio management, see Fig.\ref{fig:johnson} for a fit to error residual data. Now, if the error residual $e\sim \text{JohnsonSU}(\gamma, \delta, \xi, \lambda)$, then one can prove \cite{johnson} that 
\begin{equation}
  z=\gamma + \sinh\bigg(\frac{x-\xi}{\lambda}\bigg)\sim N(0,1),
  \label{eq:johnson_nonlinear}
\end{equation}
and thus standard outlier detection methods such as the generalized Extreme Studentized Deviate Test (gESD) \cite{gesd} can be applied.

\medskip
\noindent {\bf Nonparametric bootstrapping --} Fig.\ref{fig:P(E)} makes it clear that errors are not Gaussian as there are extreme events. At the same time, let us assume we were not able to find a parametric distribution that accurately fits the whole range. In order to extract relevant statistics from this distribution, we thus need to resort to the {\it nonparametric bootstrapping method}. This is an algorithm to accurately estimate an arbitrary  distribution statistic $x$ (e.g., the mean, a percentile, etc) from a population of $N$ observations, and is based on the following steps:

\begin{enumerate}
    \item we resample $N$ points at random from the population (with replacement, i.e. the resampling is not equal to the original population due to repetitions);
    \item we compute the statistic of interest $x$ in the new sampling;
    \item we repeat 1) and 2) sufficiently many times, to obtain a sequence of $x$’s.
    \item Since this sequence converges to a Gaussian distribution, we finally order the sequence and take the $2.5\%$ and the $97.5\%$ values, to obtain the boostrap $95\%$ confidence interval of $x$.
\end{enumerate}

For illustration, the 5th percentile of the marginal residue distribution $P(e)$ is $-0.0432$. Applying the procedure above, the bootstrap CI95 of the 5th percentile is $[-0.0353, -0.0548]$, and thus a better approximation to the true 5th percentile is $-(0.0548+0.0353)/2=-0.045$.
Similarly, the empirical 95th percentile of $P(e)$ is $0.0448$, but the bootstrap CI95 is $[0.0586, 0.0379]$, and thus a more accurate estimation is $0.048$. Therefore, we can build a prediction interval to any new prediction $\tilde{y}$ as $[\tilde{y}-0.045,\tilde{y}+0.048]$. This interval is supposed to accurately predict the true value $y$  no less than $90\%$ of the times. Section \ref{sec:uncertainty} further develops this into an uncertainty model.

\subsection{Mesoscopic error quantification: conditioning on input space}
\label{sec:error_conditioned_input}
This corresponds to box $\#6$ in Fig.\ref{fig:pipeline}.

\medskip
\noindent Whereas analysis of box $\#5$ focused on the marginalized residue distribution $P(e)$, the current subsection further inspects whether residues are homogeneously distributed across the input space or, conversely, whether there are regions, manifolds, or directions in the feature space along which the residues are disproportionally different. Finding out where errors lie in the input space and whether biases exist is useful for various reasons. First, from a practical point of view it will help us refine our understanding and calibration of the uncertainty of the ANN prediction (see section \ref{sec:uncertainty} for more details). Second, learning where error lie in input space will allow us to refine our model (e.g. by refinining the regularization function) or our database (e.g. by further augmenting the training set in the regions with high error), see section \ref{sec:boosting} for details. Finally, the large error regions can be flagging regions of the input space with strong gradients, i.e. regions which are intrinsically harder to predict.
This information can be validated by the expert stress or flight physics, thus increasing the model's interpretability (see section \ref{sec:xai}).\\
Conceptually, this current box $\#6$ is therefore about exploring possible error biases by conditioning the residue distribution on the input space, i.e., about reconstructing $P({\bf e}\vert{\bf X})$. When the number of features $n$ is large, however, an exhaustive conditioning is computationally unaffordable, and partial solutions are needed. Below we depict a set of strategies and illustrate them.

\begin{figure}[htb]
\begin{centering}
\includegraphics[width=0.95\textwidth]{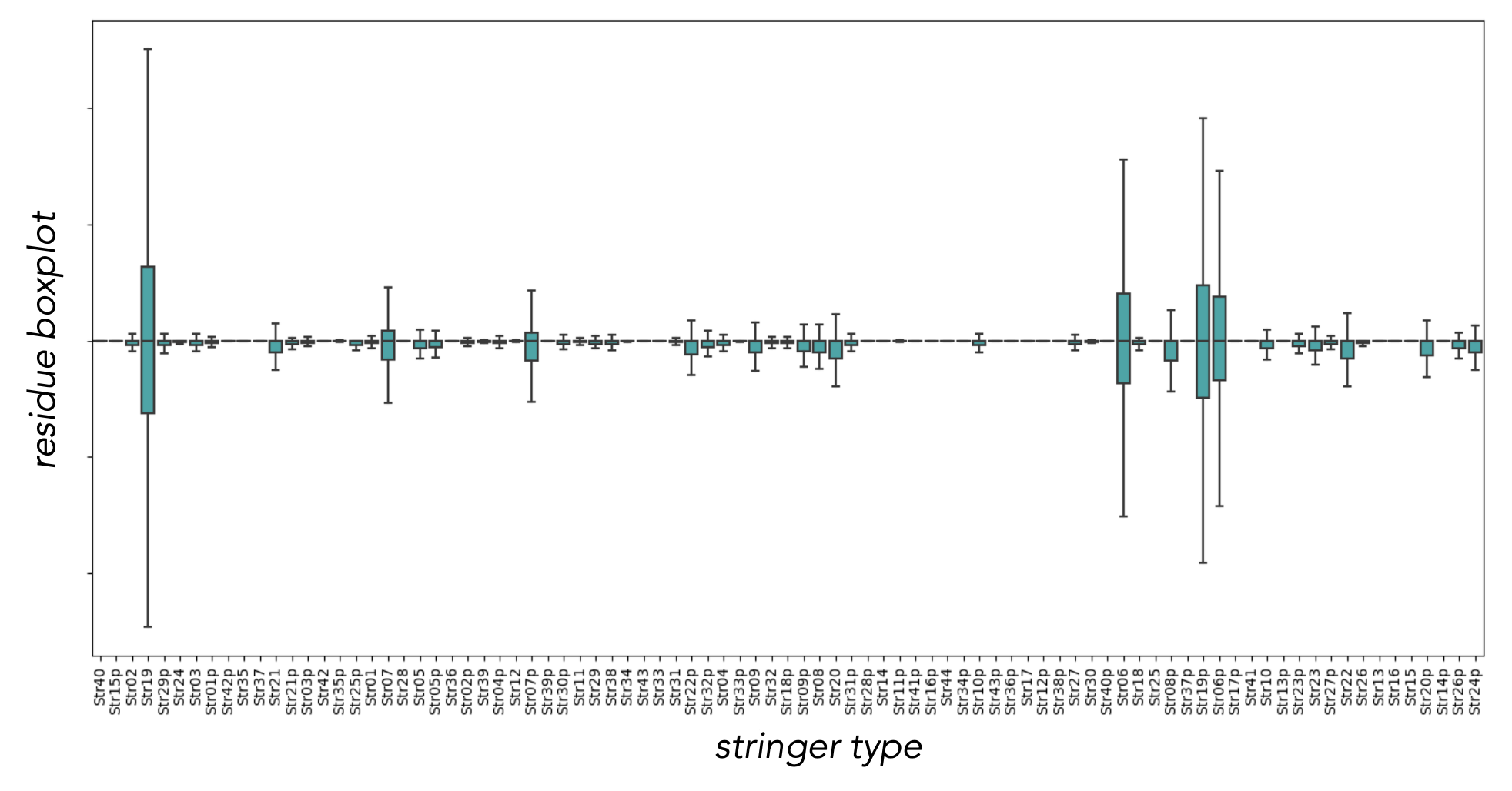}
\par\end{centering}
\caption{Residue boxplots as a function of the stringer's category, for all points in the test set. We initially filter the test set according to the stringer's category each test set point has. We can clearly see that in a few stringers, the residue variance is disproportionally larger than for the rest. This information can be used by the engineer to (i) enhance the database, or (ii) check the physics of the specific stringer outliers.}
\label{fig:stringer_input}
\end{figure}

\medskip
\noindent {\bf Conditioning on categorical variables --} A first possibility is to visualise the test set's error residuals with respect to the span of a particular categorical variable. In Fig.\ref{fig:stringer_input} we do so by conditioning on the particular stringer (a geometrical variable). We filter all test set data according to the stringer they belong to, and then make, for each stringer, a boxplot of the error residuals. Fig.\ref{fig:stringer_input} clearly shows that most stringers has similarly low residues, except for a few cases (Str19, Str06, Str19p, Str06p) where the data shows substantially more dispersion. In order to check whether this categorical variable displays some kind of bias, we perform two tests: (i) ANOVA, which analyses whether residue means across stringers are compatible, and (ii) Levene's test, which is similar to ANOVA except that the comparison between classes is not about the mean, but the variance. In this example, both tests can reject the null hypothesis with high statistical significance ($p$-value$<0.005$ in both cases).

\begin{figure}[htb]
\begin{centering}
\includegraphics[width=0.95\textwidth]{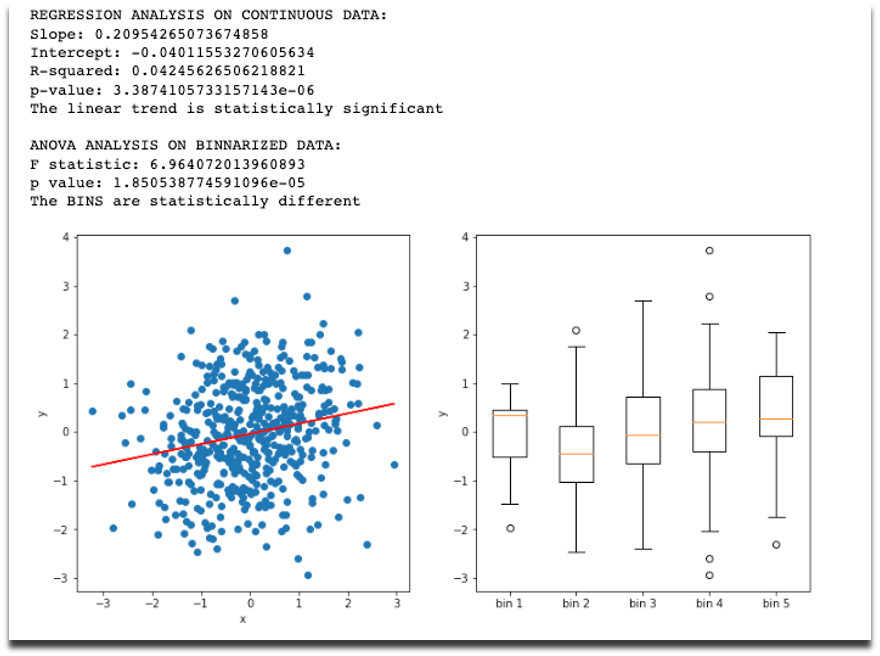}
\par\end{centering}
\caption{The left panel reports some artificial scatter plot on error residuals as a function of a numerical variable $x$, with a statistically significant fit to a linear model (Pearson correlation, left panel). In the right panel, the numerical variable $x$ has been binned into 5 bins, and an ANOVA test is performed. The null hypothesis is rejected, as expected.}
\label{fig:pearson_anova_input}
\end{figure}

\begin{figure}[htb]
\begin{centering}
\includegraphics[width=0.5\textwidth]{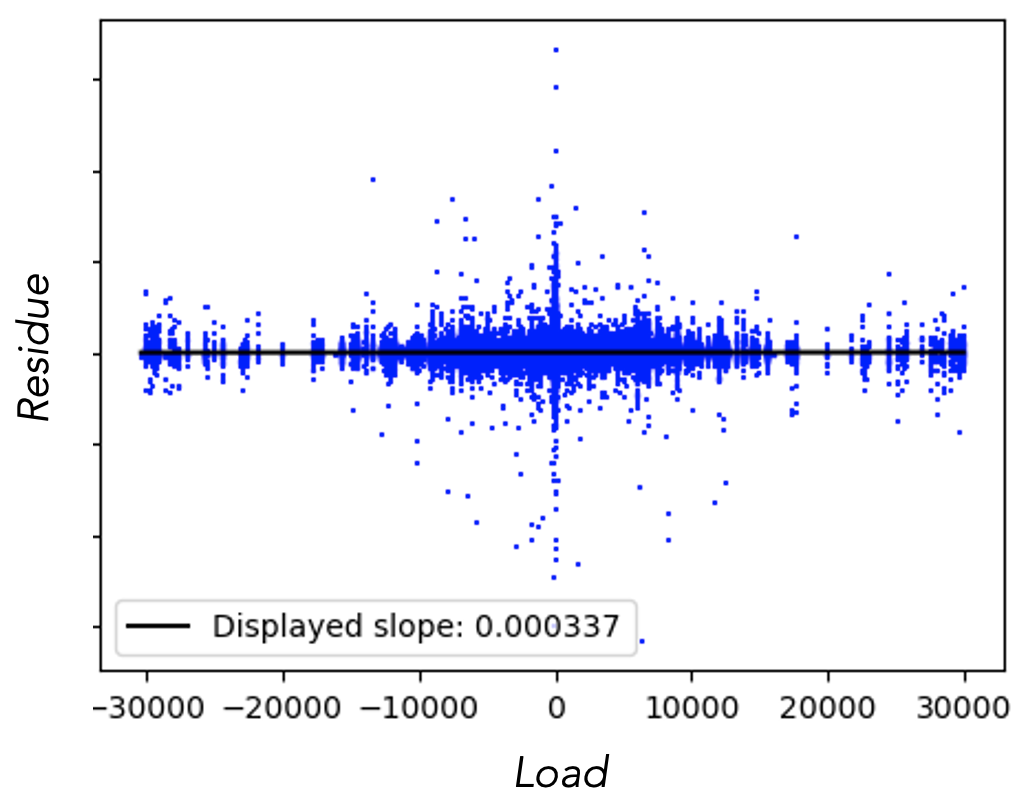}
\par\end{centering}
\caption{Error residual of the test set as a function of a particular load variable. No linear trend is displayed ($p$-value=0.59), but after binning the load variable, ANOVA's null hypothesis can be rejected ($p$-value$=0.01$).}
\label{fig:load_input}
\end{figure}

\medskip
\noindent {\bf Conditioning on numerical variables --} Conditioning on numerical variables is about plotting the error residuals of the test set as a function of a particular (numerical) variable. The goal here can be to try and establish (i) some kind of error trend (i.e. a monotonic trend) or otherwise to (ii) find some region --i.e. some interval of the numerical variable-- where error residuals are disproportionally larger. Aim (i) can be addressed by testing some linear (i.e. Pearson) or monotonic (i.e. Spearman) correlation on the data, whereas aim (ii) is better assessed by binning the numerical variable and performing an ANOVA/Levene's test. These two options are illustrated in Fig. \ref{fig:pearson_anova_input}. Observe that whereas having a linear trend (i.e. a linear bias) in some numerical variable implies rejecting the null hypothesis in an ANOVA's test on the binned variable, the converse is not necessarily true. For illustration, in Fig.\ref{fig:load_input} we plot the error residual as a function of a specific numerical variable (a load type). A linear trend on this plot is not statistically significant ($p$-value=0.59), but ANOVA test on the residual errors after binning the load variable yiels a $p$-value$=0.01$, i.e. there is a statistically significant pattern on the bin's error residual average.

\begin{figure}[htb]
\begin{centering}
\includegraphics[width=0.75\textwidth]{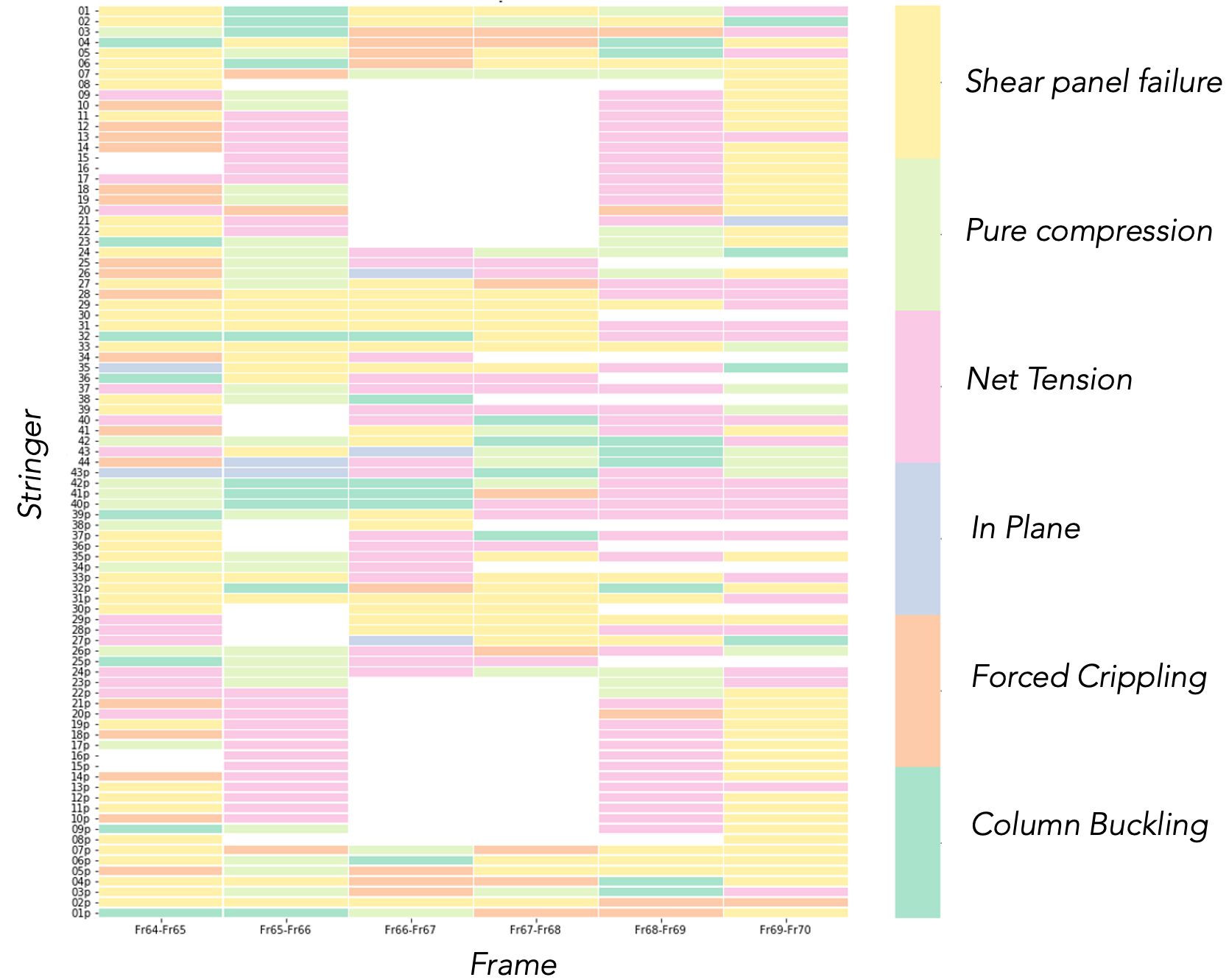}
\par\end{centering}
\caption{Stringer-Frame table, where colors correspond to the failure mode that was found to have a larger error in the precise stringer-frame duple.}
\label{fig:superstringer_residues}
\end{figure}

\medskip
\noindent The analysis above could in principle be extended to all individual input variables, both numerical and categorical. Similarly, one could condition the error residual not just on one input variable, but on a few of them, by filtering out datapoints according to the specific values they take on a subset of input variables. For illustration, in Fig.\ref{fig:superstringer_residues} we condition on the two geometrical variables (Frame and Stringer) to provide a 2D table. Each entry of the table is color-coded according to the specific failure mode that was found to harbor the largest relative error.\\
The combinatorics of these analysis quickly explode: there are $n$ individual analysis for error conditioning on a single input variable, ${n \choose 2}$ different error distributions conditioned on a pair of input variables, and so on... In general we have ${n \choose q}$ different error distributions conditioned on $q$ input variables and thus the total number of independent analysis is $\sum_{q=1}^n {n \choose q}=2^n$: a hopeless endeavor. Is there a way to know, a priori, which input variables or combination of them should be prioritised? There are two possible answers. The first approach is to use expert advice. The second is to leverage on some XAI method, as we will show in section \ref{sec:xai}.



\subsection{Mesoscopic error quantification: conditioning  on output space}
\label{sec:error_conditioned_output}
\begin{figure}[htb]
\begin{centering}
\includegraphics[width=0.9\textwidth]{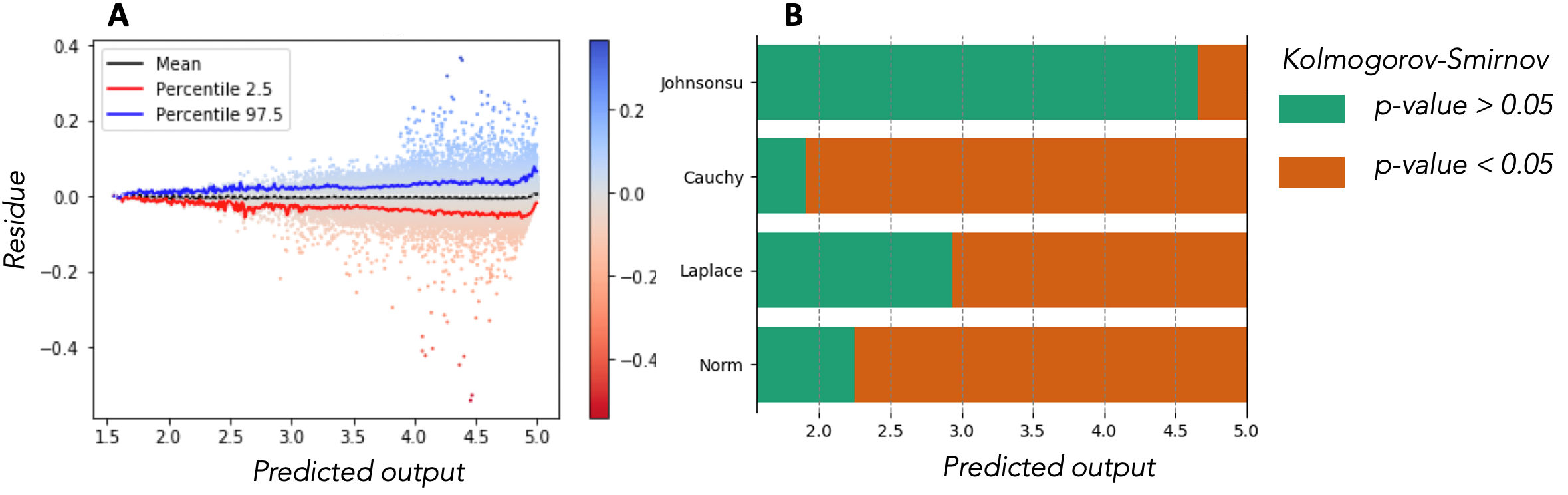}
\par\end{centering}
\caption{(A) Scatter plot of the error residuals for a given reserve factor, as a function of the value of the predicted value, showing that error is not homogeneously distributed and there are some regions in the output space where error is concentrated. (B) After binning the output variable into ten bins, we filter the test set and fit the error residuals of the filtered data inside each bin to a parametric distribution. We color-code the bin according to the $p$-value associated to the 1-sample Kolmogorov-Smirnov test between the empirical population and the parametric distribution, where we take $p$-value $< 0.05$ as a criterion for rejecting the hypothesis that data conform to that parametric distribution. }
\label{fig:superstringer_residues_2}
\end{figure}

This corresponds to box $\#7$ in Fig.\ref{fig:pipeline}.\\
Is the model performance similar for different regions of the output space? Or are there regions that concentrate more error than others? Consider, for instance, our problem of industrial interest: predicting the reserve factors of different failure modes. As previously discussed, it is much more important to be able to accurately predict the reserve factors when their actual, ground true values are small  --as this relates to a higher likelihood of failure-- than to accurately predict large reserve factors. One can start by visually inspecting this by scatter-plotting the error residuals in the test set against either the actual, ground true values $y$ or the predicted values $\tilde{y}$. An illustration is given in panel (A) of Fig.\ref{fig:superstringer_residues_2}, showing a case where we indeed find a higher variability of the residues for larger values of the reserve factor. This shows that the behavior of the residue variable is heteroscedastic. The second observation is that the uncertainty around any new prediction is {\it not} independent of the prediction itself! (according to the plot, the uncertainty is larger when the prediction gives larger values of the reserve factor). This observation will be important when building an uncertainty model (see section \ref{sec:uncertainty}). Third, not only the uncertainty but, actually, the whole error conditional residual distribution $P(e\vert y)$ could be different when conditioning for different ranges of the output variable $y$. For illustration, we have performed a binning of the output variable, filtered the test set into different bins accordingly, and fitted the (empirical) error residual distribution to a list of parametric distributions, including the Gaussian (normal) distribution, a Laplace distribution, a Cauchy distribution, and an exotic Johnson's SU distribution. We have then assessed the quality of such fit via a 1-sample Kolmogorov-Smirnov hypothesis test (where the null hypothesis is that the empirical distribution conforms to the theoretical one). Panel (B) of Fig.\ref{fig:superstringer_residues_2} shows the $p$-value of the test, for each distribution and each bin. First, we find that the Gaussian distribution is only a good fit for a very reduced interval of output variables $y<2.25$. This result is in good agreement with Fig.\ref{fig:P(E)}, except here we are making a fine-grained analysis. Similarly, the Cauchy and the Laplace distributions are only good fits in a restricted interval. The exotic Johnson's SU distribution is a better fit for a substantial interval, except for the very last bins. From this analysis, we conclude that (i) error residuals are systematically not Gaussian, (ii) for a reasonably large interval, we could fit the empirical residues to a Johnson's SU distribution, so we can use such fit for subsequent outlier detection (as in section \ref{sec:error}). Such detection of outliers can be subsequently post-processed (see section \ref{sec:boosting}).

\subsection{XAI Explainability}
\label{sec:xai}
This corresponds to box $\#8$ in Fig.\ref{fig:pipeline}.\\
How can we trust a machine learning model --such as ANNs-- if we don’t understand how and why ${\cal F}({\bf X})$ takes the values for all their parameters?
As a matter of fact, ANN training is usually seen as a black box.
{\it Explainable AI} (XAI) is an umbrella for many different strategies aimed at increasing transparency, interpretability, explainability, and eventually trust. A model is said to be {\it transparent} if the processes that extract model parameters from training data and generate labels from testing data can be described and motivated by the approach designer. On the other hand, {\it interpretability} describes the possibility of comprehending the ML model and presenting the underlying basis for decision-making in a way that is understandable to humans.
Finally, {\it explainability} refers to the collection of features (of the interpretable domain) that have contributed, for a given example, to producing a decision (e.g., classification or regression). While transparency and interpretability are abstract, ambiguous concepts, explainability is much more quantifiable. There are a number of methods in the literature e.g. SHAP (SHapley Additive Explanation) or LIME (Local Interpretable Model-agnostic Explanations), but for simplicity here we focus on one of the algorithmically simplest versions.

\begin{figure}[htb]
\begin{centering}
\includegraphics[width=0.45\textwidth]{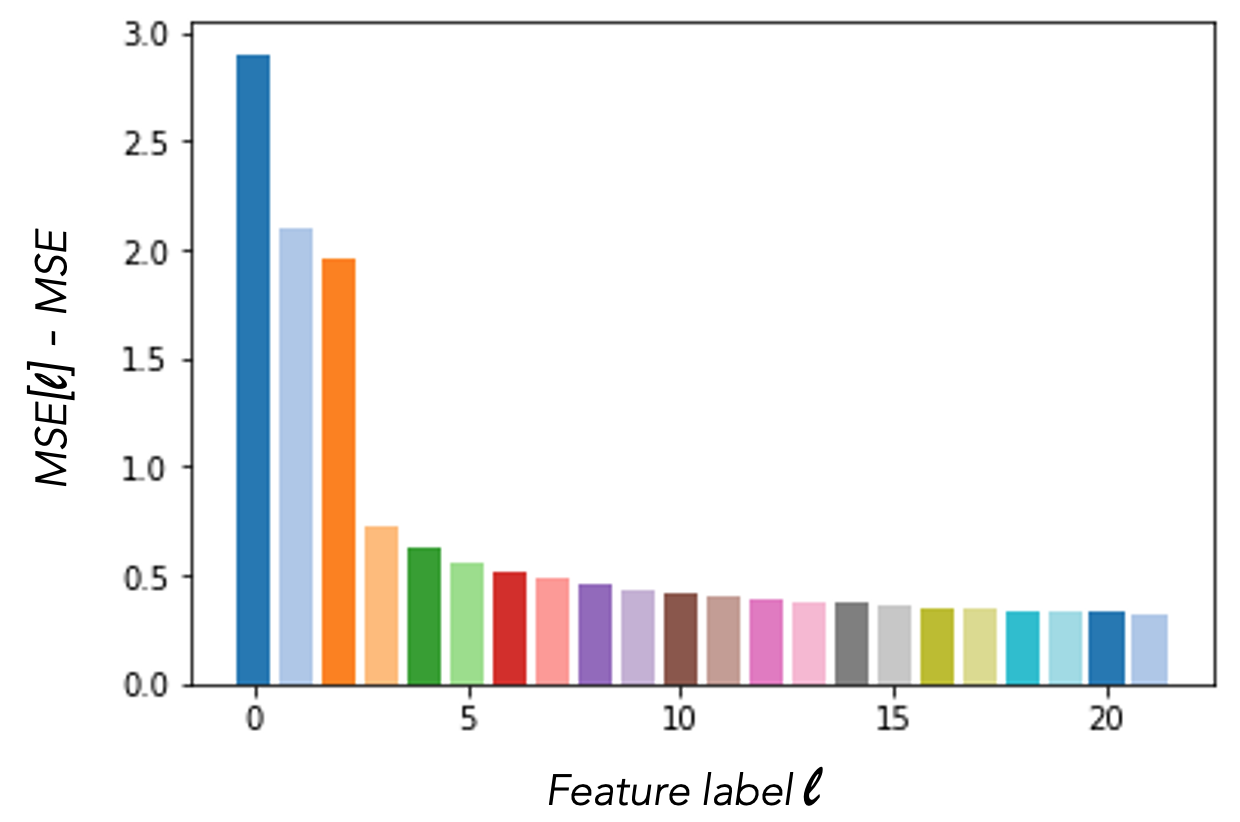}
\par\end{centering}
\caption{Illustration of the ranking induced by performing a permutation feature importance on a total of $n=22$ features, where there are three features that are clearly more important than the rest.}
\label{fig:xai}
\end{figure}

\medskip
\noindent {\bf Permutation feature importance (PFI) --} Across all the $n$ input features ${\bf X}=(x_1, x_2, x_3, \dots, x_n)$: which are the most ``important'' ones for the ANN to yield a good prediction? How much does the ANN prediction rely on each feature? The so-called {\it permutation feature importance} method assesses the impact of perturbing specific features in the prediction. It consists on the following steps:
\begin{itemize}
    \item From the test set, we construct several ``permuted'' test sets. Each permuted set is a copy of the test set where a single feature variable $x_{\ell}$ is shuffled across datapoints\footnote{Shuffling keeps the marginal distribution of $x_{\ell}$ unaltered.}. 
    \item We measure on one hand the test set error (e.g. MSE from Eq.\ref{eq:Error_MSE}), and on the other hand we measure the MSE for each of the permuted test sets $\ell=1,2,\dots,n$, where MSE[$\ell$] is the error on the set where the feature $\ell$ has been permuted. It is easy to see that $\forall \ell, \ \text{MSE}[\ell]\geq \text{MSE}$. But how larger? The larger the more important the shuffled feature is.\\
    Accordingly,
    \item We repeat this procedure for each feature $x_{\ell}$, and finally rank features according to how larger the error is with respect to the baseline error
\end{itemize}
Fig.\ref{fig:xai} provides an illustration of this method. PFI helps us to validate whether the ANN is using the information of the features as expected. For instance, the expert engineer should be able to understand and justify which are the highly ranked features. This method can be used to inform in which of the $n$ features we shall concentrate in the study referred in section \ref{sec:error_conditioned_input}.

\subsection{Model and Data Boosting}
\label{sec:boosting}
This corresponds to box $\#9$ in Fig.\ref{fig:pipeline}.
Is the model's learning architecture the correct one? Is the model `adequately' complex, or do we need to refine its hyperparameters? Is the loss function adequate, or do we need to refine the regularization terms? Are we making the most out of the model, or adding more data to the training set would improve the performance?\\
This section deals with these questions, providing a feedback loop to the overall pipeline.

\medskip
\noindent {\bf Bias and variance --} If the pointwise error metric (e.g. MSE, MAE, etc) gets similar values for the training and the test sets, then we can discard the high-variance scenario. If such values are large, then we are confronted with a high bias / low variance scenario. In this case, the model is not flexible enough to capture the nuances of the pattern. One can try to (i) increase the model's complexity (modifying the hyperparameters), or (ii) reduce the weight of the regularization term. Of course, if the error is very large then it is possible that the model's architecture itself is not sufficiently sophisticated, and one might need to upgrade it.
Conversely, if the training error is low but the test error is large, then we are in the high variance scenario: the model is perhaps `too flexible' and is learning not only the pattern but also spurious noise. To reduce its flexibility, one can try to (i) decrease the model's complexity (modifying the hyperparameters), (ii) increase the weight of the regularization term, or even (iii) try and reduce the number of input features (e.g. using feature selection, see \ref{sec:xai}). If the gap doesn't reduce after these attempts, one can also downgrade the learning architecture.

\begin{figure}[htb]
\begin{centering}
\includegraphics[width=0.85\textwidth]{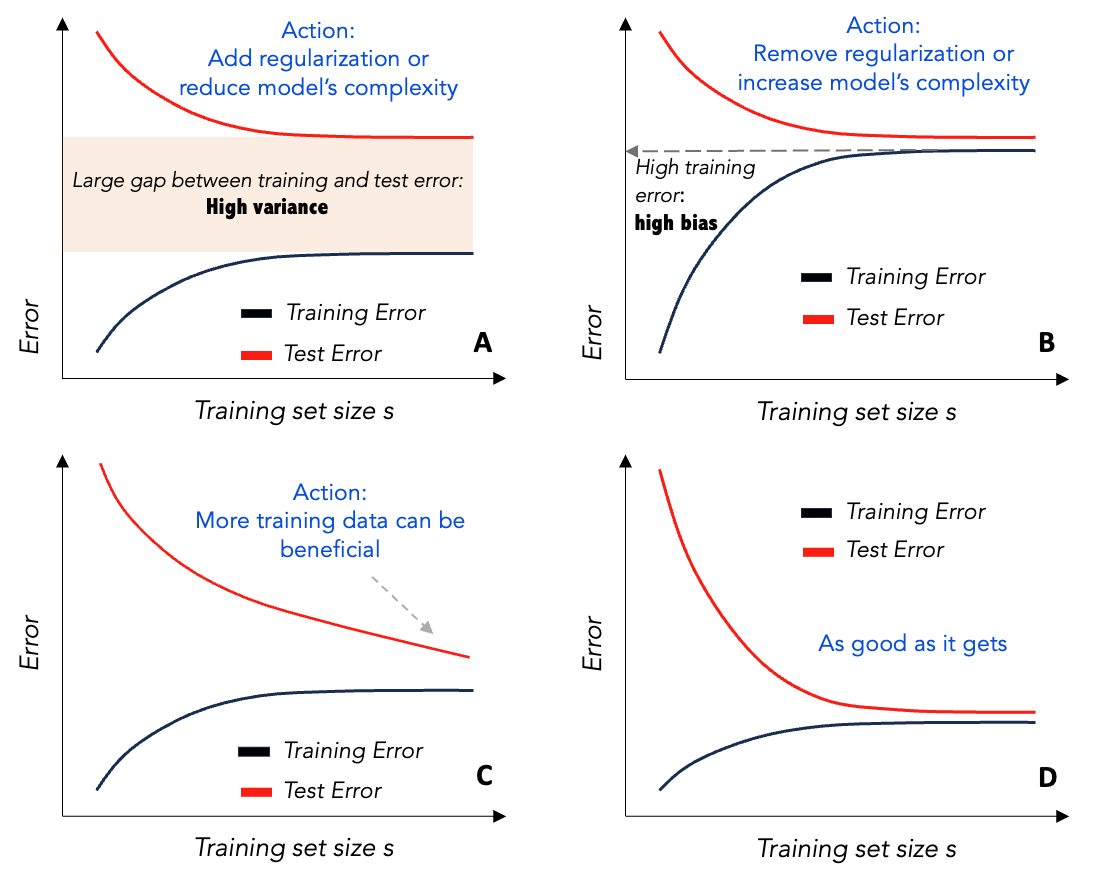}
\par\end{centering}
\caption{Learning curves of model's training and test error as a function of the training set size, covering four cases.}
\label{fig:learning_curves}
\end{figure}

\medskip
\noindent {\bf Learning curves --} 
These provide further insight about whether the model is producing high bias or high variance, and it also allows us to understand whether increasing the training set size would have a positive effect on the performance of the model. The basic idea is to fix the test set size, vary the training set size $s$, and evaluate the performance of the model (e.g. the model's error or other performance metric) both in the training set and in the test set as a function of the training set size $s$. We thus construct a total of two learning curves: training set error as a function of training set size $s$, and test set error (so-called generalization error) as a function of training set size $s$. The shape of both curves (e.g. whether curves have reached an horizontal asymptote of high or low value), and their comparison --e.g. the gap between them--, will 
allow us to identify whether the model has high or low bias and variance, to understand whether the size of our original dataset was `enough' (i.e. if the learning curves are somewhat flat when we start decimating the full dataset) or if, conversely, the model has not converged and could perform better if more data were available. A sketch of this analysis is illustrated in Fig.\ref{fig:learning_curves}. Panels (A) and (B) cover the situation where boosting the training set with more data probably won't help, as learning curves have reached plateaus. In (A), the test error plateaus at a value significantly larger than the training error, showing signs of high variance, whereas in (B) both the training and test errors plateau at very similar values, albeit these are large, pointing to high bias. Panel (C) depicts a situation where the test error has not plateaued, i.e. the model still overfits but this might be solved by boosting the dataset (adding more data) and thus pushing further down the test error towards the training error, overall increasing the model's generalization performance.
Finally, Panel (D) shows a situation where both learning reach a plateau and converge towards a low error value: this is the ideal situation where the model's performance cannot be further improved by adding more data and the model's complexity is in the sweet spot balancing bias and variance.

\medskip
\noindent {\bf Boosting dataset: data augmentation --} 
Suppose that a learning curve analysis of our model points out that the performance could be improved by increasing the size of the training dataset. How can we do this? Of course, the first option is to simply generate more data, e.g. by running numerical (CFD or structural) simulations or wind tunnel experiments. However this is often not acceptable. What to do then? We enter the realm of {\it data augmentation} \cite{augmentation, augmentation2}, which can be defined as the task of increasing the training set by suitably modifying copies of existing data. There are various ways of performing data augmentation, depending on the type of data under analysis and the type of learning task. A classical example is image classification, and under the premise that an image contains the same information (e.g. the same digit, the same type of animal, etc) under various geometric transformations or different types of noise injection, random erasing, etc \cite{augmentation}, one can generate modified copies of images with the same labels.\\
In our industrial case of study, the input data are not images, but a mix of categorical and numerical features. The numerical features (e.g. structural loads) might have some attached uncertainty. We can exploit such uncertainty for data augmentation, as explained in section \ref{sec:data}. This procedure will be specially important for those regions (e.g. the input voxels) that have been flagged as having data scarcity by the {\it voxel tesselation and proximity method} (section \ref{sec:split}). More sophisticated ways of augmenting the data include leveraging  generative models such as variational autoencoders \cite{vae,vae2} or Generative Adversarial Networks \cite{gan}, which we won't discuss here.

\medskip
\noindent {\bf Boosting dataset: feature selection --} Suppose that we need to reduce the dimensionality $n$ of the feature set. How to do this? Here we briefly discuss three strategies. First, one obvious way is to pre-process data using PCA, as explained in section \ref{sec:data}. Second, our analysis of feature importances in \ref{sec:xai} also provide a way to reduce the feature set: if the permutation feature importance test reveals that there is a subset of features for which the MSE is roughly equal to the MSE of the non-permuted set, then those features are virtually irrelevant for the learning task and one can decide to remove them. Third, by selecting a $L_1$ regularization in the loss function, sparser network configurations are rewarded. Once the network has been trained, we can now explore the weights between initial layer and the first hidden layer. Remind that the initial layer has $n$ neurons (one per feature). Accordingly, if all the weights incident to the $j$-th neuron of the first layer are very small (i.e. virtually null), then we can safely conclude that the information stored in the $j$-th neuron does not propagate further in the network, and therefore the $j$-th feature can be safely discarded.

\medskip
\noindent {\bf Boosting dataset: applicability layers --} We define the model's {\it input applicability domain} ${\cal A}_X$ as a subset of the feature set $S_X$, denoting the region of the input space for which the surrogate model can be applied. By definition, a point $\bf X=(x_1,x_2,\dots,x_n)$ in the input space can be inside the feature set ($\bf X \in S_X$) but outside the applicability domain. Once the applicability domain has been defined, then the training and test sets can be refined by removing all datapoints which are not inside ${\cal A}_X$.
Note that the applicability domain is a concept related to the classical {\it Operational Design Domain} \cite{pablo1} used in engineering to establish the conditions under which a certain product is safe to use. 
There are many ways of defining ${\cal A}_X$ --and thus one can speak of applicability layers accordingly--. For instance, applicability can be known a priori or be learned a posteriori. In the former case, the concept is strongly related to that of an Operational Design Domain. In the latter case, it can be used to boost the dataset within the iterative cycle depicted in Fig.\ref{fig:pipeline}.
Below we depict a range of different applicability layers, based on different criteria:
\begin{enumerate}
    \item  {\it Hypercube applicability}: The rationale is that a given point $\bf X$ is outside (hypercube) applicability if any of the entries $x_j$ of $\bf X$ are outside the interval spanned by the training data points $\{x_j\}$. The resulting applicability domain --sometimes defined as the applicability hypercube-- is simply constructed as the product of the intervals spanned by each input (numerical) variable. This type of applicability is very easy to check, but the criterion is rather loose (e.g. a point can be inside the applicability hypercube, and still the model might not provide a carefully controlled prediction). 
    \item {\it Convex Hull applicability}: this is a tighter criterion than hypercube applicability, and it is based on interpolability criteria (see Def.1) and seing membership of the convex hull as a necessary condition for an adequate performance of the ANN (see, however, the discussion about interpolating regime in high dimensions in sec.\ref{sec:split}). We say that $\bf X$ is outside (convex hull) applicability if $\bf X$ does not belong to $\texttt{hull}(\text{Train}_X)$. This type of applicability is computationally more expensive --as it requires deciding on the convex hull membership problem on the training set, which is usually high dimensional--. Other variations of this type of applicability include considering the CHMP not in ambient space but in some latent space, let it be PCA space, or in a latent space generated via nonlinear dimensionality reduction methods \cite{deephull}, where its computation is more efficient.
    \item {\it Error-based classifier}: the rationale here is to pre-train a classifier that detects, according to $\bf X$, those points that will have prediction error above a certain tolerance (e.g. the outliers). Such classifier is typically highly unbalanced and thus requires special techniques to be trained (i.e. the problem can be recast as an anomaly detection problem). Once the classifier is trained, any new point classified as being above the error tolerance will be flagged outside applicability.
\end{enumerate}

\begin{figure}[htb]
\begin{centering}
\includegraphics[width=0.6\textwidth]{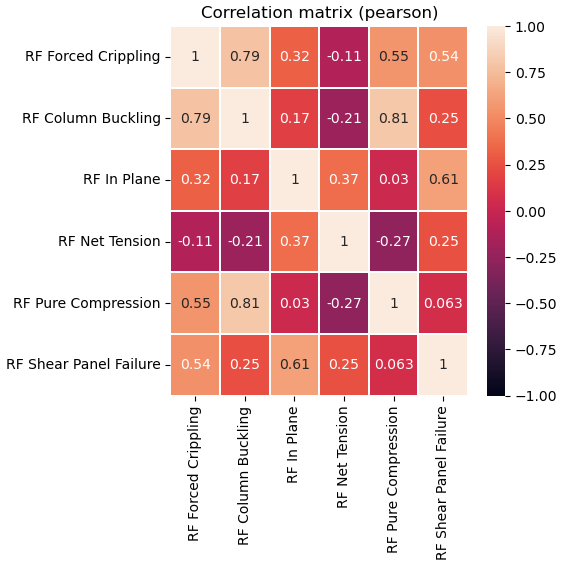}
\par\end{centering}
\caption{Heatmap illustrating of the Pearson correlation matrix $\mathscr{C}({\bf {{Y}}})$ across the $m=6$ (ground true) reserve factors, where each entry $\mathscr{C}_{ij}$ color-codes the Pearson correlation coefficient of the scatter plot between the $i-$th and $j$-th reserve factor, across all points in the training set.}
\label{fig:correlations}
\end{figure}

\medskip
\noindent {\bf Boosting the loss function --} According to our error quantification analysis (sections \ref{sec:error}, \ref{sec:error_conditioned_input}, \ref{sec:error_conditioned_output}), we might need to refine the penalization we give to certain errors. For instance, suppose we initially use the MAE as the pointwise error metric in the loss function, and we change it to:
$$\frac{1}{N}\sum_{j=1}^N \vert {\bf \tilde{Y}}(j) - {\bf Y}(j)\vert \longmapsto \frac{1}{N}\sum_{j=1}^N \vert\frac{{\bf \tilde{Y}}(j) - {\bf Y}(j)}{{\bf Y}(j)}\vert.$$
Essentially, we are changing from MAE to the so-called relative error. Observe that in this case, those errors found for smaller ground true values (which in our industrial case study mean errors found for smaller reserve factors, i.e. for higher failure likelihoods) carry more weight than errors found for large reserve factors in the computation of the loss function. Optimizing for this new loss function forces the surrogate model to being specially accurate for critical cases (those where the true reserve factor is critically low).\\
Another way in which we can boost the loss function is by refining the regularization terms. For instance, by varying the value of $\lambda$ in Eq.\ref{eq:loss} we can make the parameter regularization more or less important (recall that the error term will typically be larger if the model also needs to find a solution with low weight norm, such as the case of ridge regression). Finally, one can include additional regularization terms. For instance, in our industrial case, the output vector consists in the reserve factors of $m=6$ failure modes. Physically, these failure modes might be {\it correlated}, see Fig.\ref{fig:correlations} for an illustration. Now, the model needs to be able to ``preserve'' these correlations, as this pattern is induced by physical constraints. For one, if these correlations are preserved, somehow we only need to be able to accurately predict a handful reserve factors, as the others will be equally accurate thanks to the correlation constraints.
One can thus ``help'' the model to include such physical information by adding to the loss function a regularization term of the form $$\gamma \vert\vert \mathscr{C}({\bf Y}) - \mathscr{C}({\bf {\tilde{Y}}})\vert\vert_F,$$
where $\mathscr{C}(\bf Y)$ is the correlation matrix whose entry $\mathscr{C}_{ij}$ is the Pearson correlation of the scatterplot between the $i$-th reserve factor $y_i$ and the $j$-th reserve factor $y_j$ (for all points in the training set), $\vert\vert \cdot \vert \vert_F$ is a matrix norm, such as the Euclidean or Frobenius norm, and $\gamma$ is a parameter that tunes the importance of this term. The optimization process is now forced to (i) find good solutions (minimizing the error term), (ii) prioritizing solutions whose error for critically small reserve factors is very good (making the model accurate in those cases that are more important), (iii) minimize the norm of the model's parameters (so as to reduce overfitting), and (iv) make sure part of the physics is preserved. Deepening on this idea of preserving the physics by introducing regularization terms brings about the concept of Physics-Informed Neural Networks (PINNs) \cite{pinn}, which we won't discuss here.

\medskip
\noindent {\bf Monitoring concept drift --} In any industrial application, the database can indeed grow, making it essential to us to ensure that whatever method we choose to implement will behave properly in a near future and, even more importantly, see whether it can improve its performance with further data \cite{ecommerce}.

\subsection{Development and validation of a multiscale uncertainty model}
\label{sec:uncertainty}

This corresponds to box $\#10$ in Fig.\ref{fig:pipeline}.

\medskip
\noindent
In sections \ref{sec:error}, \ref{sec:error_conditioned_input} and \ref{sec:error_conditioned_output} we focused on the error residual's marginal distribution $P(e)$ and the conditioned distribution on the input space $P(e\vert{\bf X})$ and the output space $P(e\vert{\bf Y})$, respectively. We can combine these three analysis to build up three independent uncertainty models. Here we conceive an uncertainty model as a type of {\bf prediction interval}: given a new prediction of the $j$-th reserve factor $\tilde{y}={\mathcal F}(\bf X)\vert_j$, what interval around the prediction $[\tilde{y}-\epsilon_1, \tilde{y} + \epsilon_2]$ allows me to say that the true value $y\in [\tilde{y}-\epsilon_1, \tilde{y} + \epsilon_2]$ with $95\%$ confidence?

\begin{figure}[htb]
\begin{centering}
\includegraphics[width=0.99\textwidth]{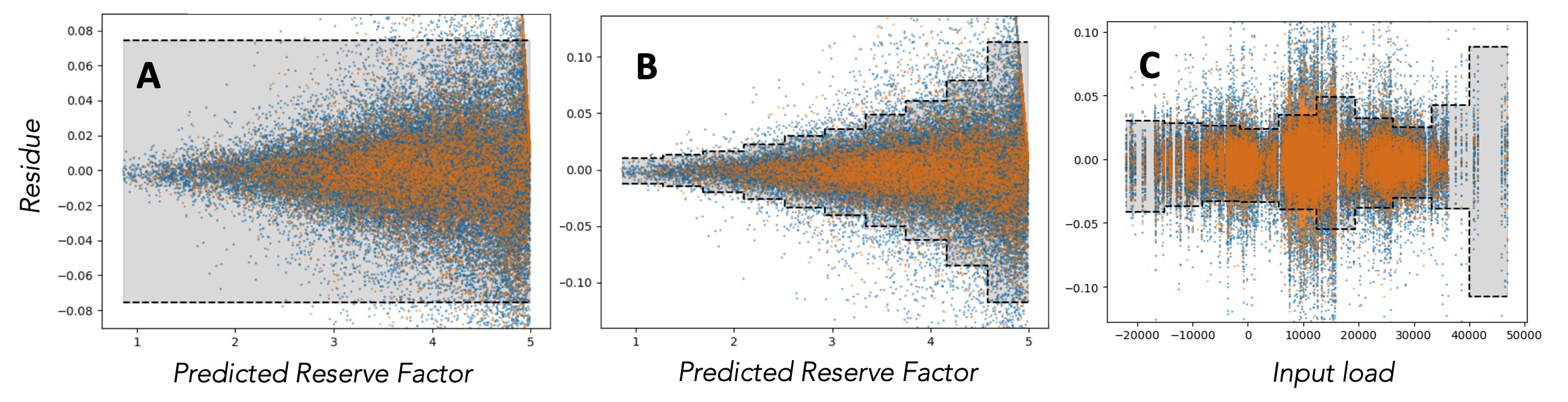}
\par\end{centering}
\caption{(A and B) Residues of all calibration set points (blue) and validation set points (orange) as a function of ground true reserve factors. The plots also depict in dashed lines (A)
 a global uncertainty model based on $P(e)$, and (B) a refined, local uncertainty model based on $P(e\vert {\tilde{y}})$. In every case the coverage is around $95\%$ (boostrap CI95 [94.63\%,95.44\%] and [94.77\%,95.56\%)], respectively. (C) Same scatter plot, but as a function of an input load $x$. The dashed lines in this case correspond to the local uncertainty model $P(e\vert x)$. Coverage of the model in the validation set is also around $95\%$ (boostrap CI95 [94.68\%,95.09\%]).}
\label{fig:uncertainty}
\end{figure}

\medskip
\noindent {\bf Global uncertainty model (GUM) --} We need to split the full dataset into training, calibration and validation sets. The ANN is trained with the training set, and $P(e)$ is computed in the calibration set (see section \ref{sec:error}). From $P(e)$, a global uncertainty model can be easily built by computing the percentiles 2.5 and 97.5 of $P(e)$ (and their boostrap confidence intervals to have more accurate estimates), so that $\epsilon_1$ is the bootstrap mean of the 2.5 percentile of $P(e)$ and  $\epsilon_2$ is the bootstrap mean of the 97.5 percentile (alternatively, we can do a more conservative model by using the bootstrap's lower bound of the CI95 of the 2.5 percentile of $P(e)$ as $\epsilon_1$, and the bootstrap's upper bound of the CI95 of the 97.5 percentile as $\epsilon_2$). For illustration, these bounds are shown in black dashed lines in Panel A of Fig.\ref{fig:uncertainty}. Finally, we validate this model by checking how many predictions in the validation set lie within the bounds of the uncertainty model. This percentage is called the {\bf error coverage}. For the model to be well calibrated, coverage needs to be larger or equal to $95\%$ (one can also compute the bootstrap CI95 of coverage, and $95\%$ needs to be inside the confidence interval).

\medskip
\noindent {\bf  Mesoscopic uncertainty models (MUMs) --} There are two type of mesoscopic uncertainty models: those based on conditioning on the input space (iMUM), i.e. percentile computation in $P(e\vert {\bf X})$ (see section \ref{sec:error_conditioned_input}), and those based on conditioning on the output space (oMUM), i.e. percentile computation in $P(e\vert {\bf Y})$ (see section \ref{sec:error_conditioned_output}).  
Examples of oMUM and iMUM are shown in Panels (B) and (C) of Fig.\ref{fig:uncertainty} for illustration.

\medskip
\noindent For each prediction $\tilde{y}$ (i.e. for each of the $m$ output variables), there's one GUM, one oMUM and $n$ iMUMs (one per input variable), i.e. $n+2$ prediction intervals. The most conservative approach is to use as the full uncertainty model the intersection interval:
\begin{equation}
    \text{FUM} = \text{GUM} \cap \text{oMUM} \cap \bigg(\bigcap_{j=1}^n \text{iMUM}[j]\bigg),
    \label{eq:FUM}
\end{equation}
although other approaches are also possible, e.g. $\text{FUM} = \text{GUM} \cap \text{oMUM}$ and discard dependencies on input space.

\section{Discussion}
\label{sec:discussion}

In this paper we have outlined and illustrated a complete model validation pipeline (integrating concepts and methods ranging from machine and deep learning to optimization and statistics) with the necessary rigor to contribute to the certification of ML-based industrial solutions \cite{codann1}. The pipeline is specifically tailored to address supervised learning models.
While the illustration of the pipeline was a realistic problem in the aeronautical industry, the pipeline is equally applicable throughout industries and serves as a recipe to assess the model design and its validation for generic industry-relevant supervised learning tasks. 
Future versions of this pipeline should be derived to deal with other learning paradigms, from reinforcement learning \cite{ML_aerospace1} to collective learning \cite{collective}. Finally, observe that the pipeline's statistical validation approach is eminently frequentist: in this sense further work should include Bayesian methods as well. \\


\noindent {\bf Acknowledgments} UPM and Airbus have collaborated in the context of a working group established by Airbus as part of the  project ONEiRE(P220021406) funded by the  Spanish Research Agency Centro para el Desarrollo Tecnológico y la Innovación (CDTI). The authors also acknowledge funding from project TIFON (PLEC2023-010251) funded by MCIN/AEI/10.13039/501100011033/, and from projects DYNDEEP (EUR2021-122007), MISLAND (PID2020-114324GB-C22), and the María de Maeztu project CEX2021-001164-M (LL) also funded by MICIU/AEI/10.13039/501100011033. \\

\end{document}


\title[Supplementary Information]{Supplementary Information of the article "What makes a network complex?"}


\author[1]{\fnm{Lucas} \sur{Lacasa}}\email{lucas@ifisc.uib-csic.es}

\author[2,3]{\fnm{Jesús} \sur{Gómez-Gardeñes}}\email{gardenes@gmail.com}

\author*[1]{\fnm{Ernesto} \sur{Estrada}}\email{estrada@ifisc.uib-csic.es}

\affil[1]{\orgdiv{Institute for Cross-Disciplinary Physics and Complex Systems (IFISC)}, \orgname{CSIC-UIB}, \orgaddress{\city{Palma de Mallorca}, \country{Spain}}}

\affil[2]{\orgdiv{Department of Condensed Matter Physics}, \orgname{University of Zaragoza}, \orgaddress{\city{Zaragoza}, \postcode{50009}, \country{Spain}}}

\affil[3]{\orgdiv{GOTHAM lab, Institute of Biocomputation and Physics of Complex Systems}, \orgname{University of Zaragoza}, \orgaddress{\city{Zaragoza}, \postcode{50018}, \country{Spain}}}


\abstract{This is the abstract}




\maketitle